\documentclass[wcp]{jmlr}

\usepackage{longtable}

\usepackage{booktabs}
\usepackage{graphicx}
\usepackage{amsmath}
\usepackage{subcaption}
\usepackage{multirow}

\usepackage{natbib}

\title[Using DNN to Automate Large Scale SA]{Using Deep Neural Networks to Automate Large Scale Statistical Analysis for Big Data Applications}

\author{\Name{Rongrong Zhang} \Email{zhan1602@purdue.edu} \\
	Department of Statistics, Purdue University\\
	West Lafayette, IN 47906 \\
	\Name{Wei Deng} \Email{deng106@purdue.edu} \\
		Department of Mathematics, Purdue University\\
			West Lafayette, IN 47906 \\
	\Name{Michael Yu Zhu} \Email{yuzhu@purdue.edu} \\
		Department of Statistics, Purdue University\\
			West Lafayette, IN 47906 \\
		}

\begin{document}

\maketitle

\begin{abstract}
	Statistical analysis (SA) is a complex process to deduce population properties from analysis of data. It usually takes a well-trained analyst to successfully perform SA, and it becomes extremely challenging to apply SA to big data applications. We propose to use deep neural networks  to automate the SA process. In particular, we propose to construct convolutional neural networks (CNNs) to perform automatic model selection and parameter estimation, two most important SA tasks. We refer to the resulting CNNs as the neural model selector and the neural model estimator, respectively, which can be properly trained using labeled data systematically generated from candidate models. Simulation study shows that both the selector and estimator demonstrate excellent performances. The idea and proposed framework can be further extended to automate the entire SA process and have the potential to revolutionize how SA is performed in big data analytics.

\end{abstract}


\section{Introduction}
According to the definitions of Gartner \cite{Gartner2014} and De Maro et al. \cite{Maro2016Bigdata}, big data refer to information assets with characteristics high volume, high velocity, and/or high variety, and the transformation from big data to value requires specific analytical methods. Currently, machine learning methods are used as the main tools for big data analytics, which emphasize algorithms instead of statistical analysis. Statistical Analysis (SA) is the process of deducing population properties and draw conclusions by analysis of data. Typically, the SA process consists of exploratory data analysis (EDA), model building, parameter estimation, hypothesis testing, interval estimation, prediction, and model diagnostics. Not only does the process depend on available computing system and software, it also heavily depends on the analyst. It takes a well-trained and experienced analyst to successfully conduct SA for data sets of conventional size. It is extremely challenging to do so for large scale data analysis. This is one of the primary reasons that SA has not been widely used for analyzing big data. 

We use multiple regression analysis as an illustrative example of the SA process. In particular, we focus on the first three steps including EDA, model building, and parameter estimation. Suppose there is a data set of one response variable $Y$ and $p$ explanatory variables $X_1, X_2, \ldots, X_p$, and it is of interest to investigate the relationship between $Y$ and $X_1, \ldots, X_p$. In the EDA step, summary statistics, empirical distributions, pairwise correlations, and scatter plots will be calculated or generated. The analyst will then need to check the values and graphics for patterns, relations, and suspicious data points. Based on the EDA results, the analyst will propose a proper regression model.  After that, statistical principles such as the least squares principle or the maximum likelihood principle are used to compute the estimates of the model parameters. 

Deep neural networks (DNNs) have achieved remarkable performances in  a variety of applications including competitive game play, object recognition, machine translation,  and speech recognition in recent years, and promise to deliver artificial intelligence (AI) in almost every aspect of human life and society. We believe that DNNs can also be used to automate the SA process, bring AI to large scale statistical analysis, and make SA popular for big data analytics. 

Due to the complexity of the SA process and space limitation, we will narrow down to two most important SA procedures, which are model building and parameter estimation,  and propose to use convolutional neural networks (CNN) 
\cite{lecun1998gradient} for their automation.  Conventionally, model building is either done by the analyst based on the results of exploratory data analysis and his/her own knowledge, or it is done via model selection using statistical principles and criteria. Both of these two approaches are however difficult to automate. Instead, we first reformulate model building as a general model selection problem, and further propose to construct a CNN to perform model selection. We refer to the resulting CNN as the {\em neural model selector}.  Data systematically simulated from candidate models will be used to train the neural model selector. 

Similarly, conventional statistical approaches for parameter estimation such as least squares estimation and maximum likelihood estimation are also difficult to automate. In this paper, we again propose to construct a CNN to perform parameter estimation.  Data systematically simulated from the underlying distribution are used to train this CNN. We refer to the resulting CNN as the {\em neural parameter estimator}. The idea of using neural networks to estimate parameters in a stochastic model is not entirely new, but it is neither well-known nor a common practice in the statistical community. We will give a brief literature review of this idea in the next section.

We further explore the possible interplay between the neural model selector and the neural parameter estimator. As will be discussed in the Proposed Approaches section, the two CNNs can be entirely separate, almost identical, or partially joint, leading to different performances in training as well as in application. We carry out extensive simulation studies, and show that the proposed neural model selector and parameter estimator can be properly trained, and the trained CNNs demonstrate excellent performance in test data and a real data application. The idea and proposed framework can be further extended to the entire SA process with the potential to change how SA is done in conventional data analysis and big data analytics. 

\section{Related work}

There exists an extensive statistical literature on model selection \cite{bozdogan1987model, burnham2003model, burnham2004multimodel}. Numerous model selection methods have been proposed. Some of these methods are not applicable to the setting we consider in this paper, while others, though applicable, may run into various difficulties;  see discussions in Section \ref{sec:proposedApproach} for details. To our best knowledge, there is no prior work about redefining the model selection problem as a machine learning classification problem and training CNN to learn and perform model selection with labeled simulated data. 

There also exist a variety of statistical methods for parameter estimation in the literature; see \cite{casella2002statistical, huber1964robust,norton2010identification}. Most statistical methods rely on full or partial knowledge of the model and are based on statistical principles. After conducting intensive literature search, we only found one paper \cite{xie2007estimation}, in which the authors proposed to use artificial neural networks and simulated data to construct estimates for parameters of a stochastic differential equation. However, the idea of using CNNs and simulated data to automate parameter estimation and model selection and bring AI to the general SA process appears to be novel to our best knowledge. 

\section{Proposed approach}\label{sec:proposedApproach}

As discussed in the Introduction, we first reformulate model building and parameter estimation as a machine learning problem. Suppose $\mathcal{M}=\{M_k: 1\le k \le K\}$ be a collection of $K$ prespecified models/distributions. Let $f(y|\theta_k, M_k)$ be the density function of model $M_k$, where $\theta_k\in \Theta_k$ is the scalar parameter of the density function. Assume that we have a random sample from one of the models, which is $\{y_j\}_{1\le j \le N}$, but we do not know the data-generating model and its parameter. The goal of statistical analysis is to identify the model and further estimate the model parameter. 

To achieve the analysis goal stated above, conventionally, the statistician will employ various model selection methods together with some estimation methods. Here, we will briefly discuss several representative approaches, which include the Kolmogorov-Smirnov (KS) distance \cite{chakravarti1967handbook}, Bayesian Information Criterion (BIC) \cite{schwarz1978}, and Bayes factor \cite{kass1995bayes}.  The KS distance method calculates the distance between the population Cumulative Distribution Function (CDF) and the empirical CDF based on the sample $\{y_j\}$ for each model. The model that achieves the minimum distance will be selected as the true model. The BIC criterion calculates the BIC score for each model as follows:
\[
\mbox{BIC}(M_k)=-2\mbox{log} L(\hat{\theta}_k) + \mbox{log} (n) p
\]
where $L(\cdot)$ is the likelihood function, $\hat{\theta}_k$ is the maximum likelihood estimate, and  $p$ is the number 
of parameters in the model $M_k$. Note that for the scenario considered here, $p=1$.  The model that achieves the minimum BIC score will be selected as the true model. 

The Bayes factor method will impose a prior distribution to the models, $\pi(M_k)$,  and further impose a prior distribution to the parameter $\pi(\theta_k)$. Then, given the sample, the posterior distribution for each model can be calculated, which is denoted as $\pi(M_k|\{y_j\})$.  The Bayes factor between any two models, $M_{k_1}$ and $M_{k_2}$, can be calculated as $\mbox{BF}(M_{k_1}, M_{k_2})=\pi(M_{k_1}|\{y_j\})/\pi(M_{k_2}|\{y_j\}),$  which can be used to discriminate between the two models. The model the BF scores support the most will be selected as the true model.

Our criticism for the conventional statistical approaches discussed above is two-fold. First, for the goal of automating model selection, the model set $\mathcal{M}$ usually consists of a large number of  candidate models, and the models are of huge variety. The conventional statistical methods such as the KS distance and BIC only work for selection between nested or other well-structured models. Second, for a given sample, all the conventional methods will have to calculate a score for each of  the candidate models, and then compare them to pick the winner. This can become computationally intensive or even intractable, especially for the Bayes factor approach.  Similar discussion and criticism can be applied to using conventional statistical methods for automating parameter estimation, which  we omit them due to space limitation. 

In this section, we instead propose to use CNNs and machine learning to automate model selection and parameter estimation. Our main idea is that the procedures for model selection and parameter estimation can be considered  mappings from the sample 
to a model and a value of the model parameter, that is, 
\[
G: \{y_j\}    \rightarrow  \left(\begin{array}{c} G_1(\{y_j\}) \\ G_2(\{y_j\}) \end{array}\right) \in \mathcal{M}\times \Theta
\]
where $G=(G_1, G_2)$ consists of the model selection mapping $G_1$ and the parameter estimation mapping $G_2$, and $\Theta$ is the parameter space.  Instead of using statistical principles to derive  $G_1$ and $G_2$, we propose to use CNNs to approximate them.  From here on in the rest of the paper, we refer to $G_1$ as the neural model selector, and $G_2$ the neural parameter estimator, as discussed in the Introduction.

\subsection{Labeled data and loss functions}\label{subsec:labeleddataGenerating}
As an analogy, the sample $\{y_j\}$ can be considered an image of an object, $G_1$ the classifier for object recognition, and $G_2$ the regression procedure for object localization in image processing \cite{girshick2014rich}. In order to train $G_1$ and $G_2$, just like in image processing, labeled data must be available. We propose to generate the labeled data as follows.

Let $N$ be a prespecified sample size. For each model $M_k$, we first place an equally space grid over the parameter space $\Theta_k$, which is $\{\theta_{k,1}, \theta_{k,2}, \cdots, \theta_{k,n_k}\}$. For each value of the grid $\theta_{k,l}$,  we generate $D$ samples of size $N$ from $f(y|\theta_{k,l}, M_k)$. We denote the samples as $\{y_{(k,l,d)j}\}_{1\leq j \leq N}$ for $1\le k \le K$, $1\leq l \leq n_k$, and $1\leq d \leq D$. In total, we have the following collection of labeled data 
$
Y=\left\{ \left(\{y_{(k,l,d)j}\}_{1\leq j \leq N}, M_k, \theta_{k,l}\right)_{k,l,d}\right\},
$ 
and $Y$ will be used to train and validate both $G_1$ and $G_2$. 

In order to train $G_1$ and $G_2$, we need to choose proper loss functions. As mentioned previously, the neural model selector is essentially a classifier and similar to the classifier for object recognition. Therefore, we choose the commonly used softmax loss function for training $G_1$. For details of the softmax loss function, the reader is referred to \cite{bishop2006pattern}. The neural parameter estimator $G_2$  is essentially a regression function and similar to the regression CNN for localizing an object in an image. For object localization, the $L_2$ loss is typically used, which is $L(G_2, \theta)=\| G_2-\theta\|_{2}^2$.  The $L_2$ loss function works well for image processing. For the neural parameter estimator, the $L_2$ loss is sensitive to extreme observations generated from models with long tails in $\mathcal{M}$, which makes the training process unstable. Resolve this issue, the Huber loss \cite{huber1964robust} is employed in training of neural estimator to improve the robustness against outliers. The Huber loss is defined as follows:
\begin{equation*}
L_{\delta}(\theta, \hat{\theta})=
\begin{cases}
\frac{1}{2}(\theta-\hat{\theta})^2 \quad\quad \quad   \text{ for } |\theta-\hat{\theta}|\leq \delta,\\
\delta(|\theta-\hat{\theta}|-\frac{1}{2}\delta) \quad  \text{otherwise.} \\
\end{cases}
\end{equation*}

\subsection{Two types of architectures}\label{subsec:Architecture}

The last important issue is about the architectures of the neural model selector and parameter estimator. There are two different types of architectures involved. The first type is regarding the architectures of the CNNs, which are about the numbers and sizes of the covolutional and fully connected  layers. We refer to this type of architecture as the CNN architecture. The second type of architectures is regarding the interplay between the model selector $G_1$ and the parameter estimator $G_2$. Because this type of architecture directly affects how the overall analysis performed, we refer to it as the SA architecture. 

\begin{figure}[!h]
	\centering
	\includegraphics[scale=0.5]{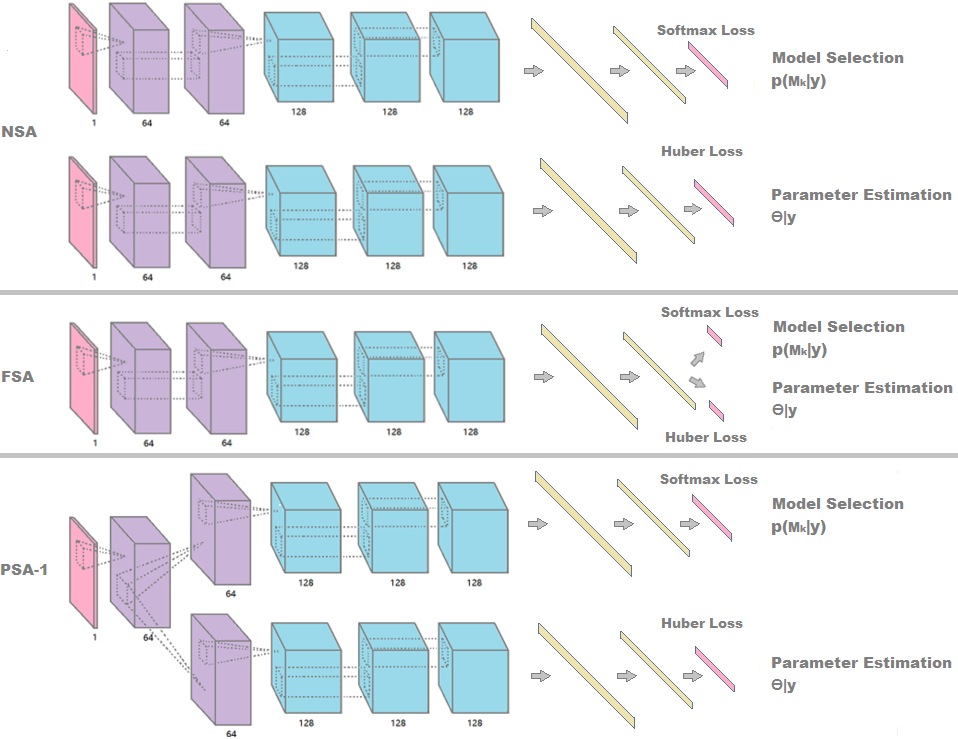}
	\caption{SA Architecture, from top to bottom: NSA, FSA, PSA-1}
	\label{fig: SA Architecture}
\end{figure}

There are three possible SA architectures. The first SA architecture uses two separate CNNs for the model selector and the parameter estimator, respectively, which we refer to as the Non-Shared Architecture (NSA). The second SA architecture uses one single CNN for both $G_1$ and $G_2$, and they part their ways only at the output layer. We refer to this architecture as the Fully Shared Architecture (FSA). The third architecture uses two partially joint CNNs for $G_1$ and $G_2$, respectively. The two CNNs can share from one to all common convolutional and fully connected layers. We refer to this architecture as the Partially Shared Architecture (PSA). The PSA sharing $l$ layers is denoted as PSA-$l$. The three SA architectures NSA, FSA, and PSA are illustrated in Figure \ref{fig: SA Architecture}.

NSA, FSA, and PSA have their own advantages and disadvantages. Using again the analogy of object recognition and localization, the convolutional layers are used to learn features that can be efficiently and effectively used for identifying the true model and its parameter.  When NSA is used, the two separate CNNs are learning the features for model selection and parameter estimation, separately, and this simple SA architecture allows easy implementation of the training algorithms. The disadvantage of NSA is that it uses only the marginal distribution of model label and true parameter value separately, instead of the entire information in their joint distribution.

When FSA is used,  the single CNN is trying to learn the same set of features, and hope that they can be used to not only select the model correctly but also estimate the parameter accurately. This is based on the assumption that such a set of common features exists. This assumption holds for distributions that belong to the Exponential family,  and minimal sufficient statistics can serve as the set of common features. This assumption however may not hold in general. Therefore, FSA is expected to work well under one set of candidate models, but may fail under another set of candidate models.

The most promising architecture is PSA. The intuition underlying PSA is that the early convolutional layers will produce low-level features that are common for both model selection and parameter estimation, and information in the true model label and parameter values can be shared. Because model selection and parameter estimation are two different tasks, we should not expect they would be relaying on the same set of high-level features. Our simulation studies reported in later sections support this intuition. In terms of training, PSA is more demanding than the other two architectures. Furthermore, PSA leads to another important issue, that is, how many convolutional layers should be shared by $G_1$ and $G_2$. We will investigate this issue in the next section.

\section{Simulation results}

We conduct simulation studies to demonstrate the properties and performance of the proposed model selector and 
parameter estimator in this section, and further compare them with several conventional statistical methods. Due to space 
limitation, we will emphasize on results regarding model selection instead of parameter estimation. The latter can be found in the Supplementary Document. 

\subsection{Setup and datasets}

The difficulty of model selection and parameter estimation depends on the number of candidate models ($K$) in $\mathcal{M}$. We consider three levels of $K$, which are $5, 20$ and $50$. We take 50 probability distributions from the textbook \cite{casella2002statistical} and some R packages \cite{smoothmest,RMTstat,LaplacesDemon,rmutil,VGAM}. The 50 distributions are listed in Table S1 in the Supplementary Document. When $K=5$,  the set of candidate models $\mathcal{M}$ includes the top five distributions in the list; when $K=20$, $\mathcal{M}$ includes the top 20 distributions in the list; and when $K=50$, $\mathcal{M}$ includes all the 50 distributions. 

The performance of the proposed model selector and parameter estimator depends on the sample size. We consider three levels of the sample size $N$, which are $100, 400$, and $900$. Each sample will be resized to a square matrix to feed into CNNs. According to the data-generating scheme described in Section \ref{subsec:labeleddataGenerating}, the total amount of training samples further depends on the number of parameter values on the grid ($n_k$) and the number of replicated samples ($D$). We specify $D=1000$, and grid size $n_k$ is set to be between 10 and 12.  For each distribution, if the parameter space is bounded, like probability $p$ in Bernoulli distribution, we will place the grid on the original space. If the parameter space is not bounded, like $\mu$ in Normal distribution, we will set the parameter space to be a bounded interval, $[0, 12]$. Following the scheme of Section \ref{subsec:labeleddataGenerating}, we generate all the labeled data, 80\% of which is used for training, and the other 20\% is used for validation. Note that we use the definitions given in \cite{ripley2007pattern}: a training set is used for learning, a validation set is used to tune the network parameters, and a test set is used only to assess the performance of the network. 

We further generate the test data as follows. For each model, we first randomly sample 100 parameter values from the parameter space, those values are just in the same range as parameters in the training set, but not the same; and second, for each sampled parameter value, we generate 10 random samples of size $N$. We test the trained model selector and parameter estimator using the test datasets. Counting both the labeled data and test data, in total, we have generated roughly 400 thousand training samples, 100 thousand validation samples,  and 50 thousand test samples for the 50 models.

\subsection{Architecture setup and training}

In Section  \ref{subsec:Architecture}, we discussed the three possible SA architectures (NSA, FSA, and PSA), but have not discussed the CNN architectures. In our simulation studies, we employ three different sizes of CNNs, which are referred to as small, medium, and large, respectively. The small CNN architecture consists of three convolutional layers with 64, 128 and 128 filters, respectively, which are followed by two fully-connected layers with 512 and 256 neurons, respectively. The medium CNN architecture consists of five convolutional layers with 64 filters each, which are followed by two fully connected layers each with 64 neurons.  The large CNN architecture consists of five convolutional layers with 64, 64, 128, 128 and 128 filters, respectively, which are  followed by two fully connected layers with 1024 and 512 neurons, respectively. In all three CNN architectures,  convolutional filters are connected to a $5\times5$ region of their input, $2\times2$ max pooling and $2\times2$ average pooling are performed between some consecutive convolutional layers. The same CNN architecture is used for both the neural model selector and parameter estimator except for the output layers. 

Under each combination of SA architectures (NSA, FSA,  PSA), CNN architectures (small, medium, large), number of candidate models ($K=5, 20, 50$), sample sizes ($N=100, 400, 900$), we use the generated labeled data to train, validate, and test the proposed neural selector and parameter estimator. For PSA, we further vary the number of shared layers ($l$) in training. Each training run is replicated six times to assess the stability of the training procedure and results. All training is performed using the Caffe implementation \cite{jia2014caffe} on one GTX-1080 GPU. The running time each training run takes ranges from five minutes to one hour depending on the values of $K$ and $N$.

\subsection{ Performance of neural selector and estimator}

Due to space limitation, we report more detailed results of our simulation studies in the Supplementary Document and only select part of them to report in this section. Overall, the trained model selector and parameter estimator demonstrate excellent performances. 

\vspace{0.1in}

{\em Accuracy of the model selector}  Table \ref{class_accuracy} presents the performance of the model selector on the test dataset under all the combinations of SA architecture, CNN architecture, number of candidate models $K$, and sample size $N$. The selection accuracy together with standard deviation in  parentheses based on repeated six runs are reported, the better result between NSA and PSA SA architectures is denoted as bold. For PSA, we report the best results based on layer analysis, they are PSA-3, PSA-2, and PSA-5 for small, medium, and large CNN architectures, respectively. 

The table shows that the selection accuracy decreases as $K$ increases under fixed $N$, and the accuracy increases as we have larger sample size. When we have a moderate sample size, 400, the partially shared CNN architecture can achieve more than 90\% selection accuracy. In order to maintain high accuracy for large number of candidate models, large sample sizes should be used. 

Figure \ref{confusionmatrix} in Supplementary Document shows the confusion matrix based on large CNN and PSA-5 neural model selector on test dataset with $K=20$ distributions. The performance of the parameter estimator on the test dataset under different scenarios is reported in Table \ref{estimate_accuracy}. Figure \ref{scatter1}, \ref{scatterplotpart1}-\ref{scatterplotpart5} in Supplementary Document show the scatter plots of true parameter values and predicted values estimated by parameter estimator under different architectures. Overall, the large CNN PSA-5 parameter estimator performs the best.

\begin{table}[ht!]
	\caption{Model selection results under all the combinations of SA architecture, CNN architecture, number of candidate models $K$, and sample size $N$. }
	\begin{center}
		\begin{tabular}{cccccccc}
			\toprule
			& \multirow{2}{*}{\small{CNN architecture}} & \multicolumn{2}{c}{\small{$N=100$}} & \multicolumn{2}{c}{\small{$N=400$}} & \multicolumn{2}{c}{\small{$N=900$}} \\  \cmidrule{3-8}
			& & \small{NSA} & \small{PSA} & \small{NSA} & \small{PSA} & \small{NSA} & \small{PSA} \\
			\midrule
			\multirow{6}{*}{\small {$K=5$}} & \multirow{2}{*}{small } & 96.88\% & \textbf{96.92\%} &  97.68\% & \textbf{97.78\%}  & 97.98\% & \textbf{98.01\%} \\ 
			&                                                   &  (0.12\%) & (0.11\%)  & (0.19\%) & (0.13\%) & (0.13\%) & (0.07\%) \\ \cmidrule{3-8}
			& \multirow{2}{*}{medium } & 96.06\% & \textbf{96.62\%} &  97.68\% & \textbf{97.93\%}  & 97.85\% & \textbf{97.90\%} \\
			&                                                    & (0.29\%) & (0.21\%)  & (0.25\%) & (0.16\%) & (0.08\%) & (0.16\%) \\ \cmidrule{3-8}
			&  \multirow{2}{*}{large }  & \textbf{97.30\%} & 97.13\% & \textbf{97.88\%} & 97.77\% & 98.01\% & 98.01\% \\ 
			&                                                  & (0.19\%) & (0.13\%) & (0.08\%) & (0.08\%) & (0.13\%) & (0.17\%) \\ 
			\midrule
			\multirow{6}{*}{\small {$K=20$}} & \multirow{2}{*}{small }  & 90.76\% & \textbf{91.44\%} &  96.34\% & \textbf{96.59\%}  & 97.79\% & \textbf{97.81\%} \\ 
			&                                                    & (0.17\%) & (0.20\%) & (0.33\%) & (0.27\%) & (0.20\%) & (0.24\%) \\ \cmidrule{3-8}
			& \multirow{2}{*}{medium }  & 67.73\% & \textbf{88.98\%} &  95.11\% & \textbf{96.07\%}  & 97.78\% & \textbf{98.03\%} \\ 
			&                                                    & (3.09\%) & (0.86\%) & (0.47\%) & (0.46\%) & (0.14\%) & (0.08\%) \\ \cmidrule{3-8}
			&  \multirow{2}{*}{large }   & 92.18\% & \textbf{92.53\%} & 97.09\% & \textbf{97.19\%} & 98.37\% & \textbf{98.47\%} \\ 
			&                                                   & (0.23\%) & (0.35\%) & (0.23\%) & (0.32\%) & (0.29\%) & (0.14\%) \\ 
			\midrule
			\multirow{6}{*}{\small {$K=50$}} & \multirow{2}{*}{small } & 73.33\% & \textbf{74.00\%} & 86.34\% & \textbf{86.52\%}  & \textbf{90.10\%} & 90.00\% \\ 
			&                                                   &  (0.54\%) & (0.88\%) & (0.29\%) & (0.59\%) & (0.59\%) & (0.35\%) \\ \cmidrule{3-8}
			& \multirow{2}{*}{medium }  & 48.93\% & \textbf{71.61\%} & 83.86\% & \textbf{87.21\%}  & 88.72\% & \textbf{90.38\%} \\ 
			&                                                   &  (1.92\%) & (0.66\%) & (0.37\%) & (0.41\%) & (0.25\%) & (0.34\%) \\ \cmidrule{3-8}
			&  \multirow{2}{*}{large }   & 75.77\% & \textbf{78.19\%} & 88.58\% & \textbf{88.98\%} & 91.08\% & \textbf{91.11\%} \\ 
			&                                                   & (0.64\%) & (0.48\%) & (0.50\%) & (0.31\%) & (0.28\%) & (0.42\%) \\ 
			\bottomrule
		\end{tabular}
	\end{center}
	\label{class_accuracy}
\end{table}

\vspace{0.1in}

\begin{table}[ht!]
	\caption{Parameter estimation results under all the combinations of SA architecture, CNN architecture, number of candidate models $K$, and sample size $N$. The Huber Loss with standard deviation in parentheses based on six repeated runs are reported,  the better result between NSA and PSA SA architectures is denoted as bold. For PSA, we report the best results based on layer analysis, they are PSA-3, PSA-2, and PSA-5 for small, medium, and large CNN architectures respectively. We can see that PSA performs better than NSA in most cases.}
	\begin{center}
		\begin{tabular}{cccccccc}
			\toprule
			& \multirow{2}{*}{\small{CNN architecture}} & \multicolumn{2}{c}{\small{$N=100$}} & \multicolumn{2}{c}{\small{$N=400$}} & \multicolumn{2}{c}{\small{ $N=900$ }} \\  \cmidrule{3-8}
			
			& & \small{NSA} & \small{PSA} & \small{NSA} & \small{PSA} & \small{NSA} & \small{PSA} \\
			\midrule
			\multirow{6}{*}{\small {$K=5$}}  & \multirow{2}{*}{small }  & 0.074 & \textbf{0.072} &  0.022 & 0.022  & \textbf{0.014} & 0.015 \\ 
			&                                                     &  (0.0024) & (0.0041) & (0.0008) & (0.0004) & (0.0007) & (0.0003) \\ \cmidrule{3-8}
			
			& \multirow{2}{*}{medium } & \textbf{0.071} & 0.072 &  0.020 & 0.020  & 0.012 & 0.012 \\ 
			&                                                   &   (0.0011) & (0.0032) & (0.0003) & (0.0003) & (0.0003) & (0.0002) \\ \cmidrule{3-8}  
			&  \multirow{2}{*}{large }  & 0.068 & \textbf{0.067} & 0.019 & \textbf{0.018} & \textbf{0.011} & 0.012 \\ 
			&                                                   & (0.0022) & (0.0020) & (0.0002) & (0.0003) & (0.0002) & (0.0001) \\ 
			\midrule
			\multirow{6}{*}{\small {$K=20$}} & \multirow{2}{*}{small }  & 0.887 & \textbf{0.830} &  0.285 & \textbf{0.243}  & 0.154 & \textbf{0.130} \\ 
			&                                                   & (0.0310) & (0.0261) & (0.0097) & (0.0090) & (0.0106) & (0.0077) \\ \cmidrule{3-8} 
			& \multirow{2}{*}{medium }  & \textbf{0.851} & 0.864 &  0.223 & \textbf{0.207}  & 0.111 & \textbf{0.102} \\ 
			&                                                  & (0.0274) & (0.0216) & (0.0105) & (0.0103) & (0.0043) & (0.0032) \\ \cmidrule{3-8}
			&  \multirow{2}{*}{large }  &  0.752 & \textbf{0.732} & 0.203 & \textbf{0.187} & 0.110 & \textbf{0.099} \\ 
			&                                              & (0.0472) & (0.0431) & (0.0076) & (0.0043) & (0.0045) & (0.0039) \\ 
			\midrule
			\multirow{6}{*}{\small {$K=50$}} & \multirow{2}{*}{small } &  2.056 & \textbf{1.734} & 0.91 & \textbf{0.636}  & 0.654 & \textbf{0.428} \\ 
			&                                                   & (0.0724) & (0.0344) & (0.1323) & (0.0137) & (0.0483) & (0.0176) \\ \cmidrule{3-8}
			& \multirow{2}{*}{medium } &1.691 & \textbf{1.596} & 0.561 & \textbf{0.484}  & 0.313 & \textbf{0.282} \\ 
			&                                                   &  (0.0314) & (0.0521) & (0.0169) & (0.0153) & (0.0049) & (0.0132) \\ \cmidrule{3-8}
			&  \multirow{2}{*}{large }  & 1.472 & \textbf{1.258} & 0.480 & \textbf{0.399} & 0.288 & \textbf{0.246} \\ 
			&                                                 & (0.0505) & (0.0273) & (0.0114) & (0.0062) & (0.0313) & (0.0051) \\ 
			\bottomrule
		\end{tabular}
	\end{center}
	\label{estimate_accuracy}
\end{table}

\begin{figure}[ht!]
	\centering
	\subfigure[]{{\includegraphics[width=6cm]{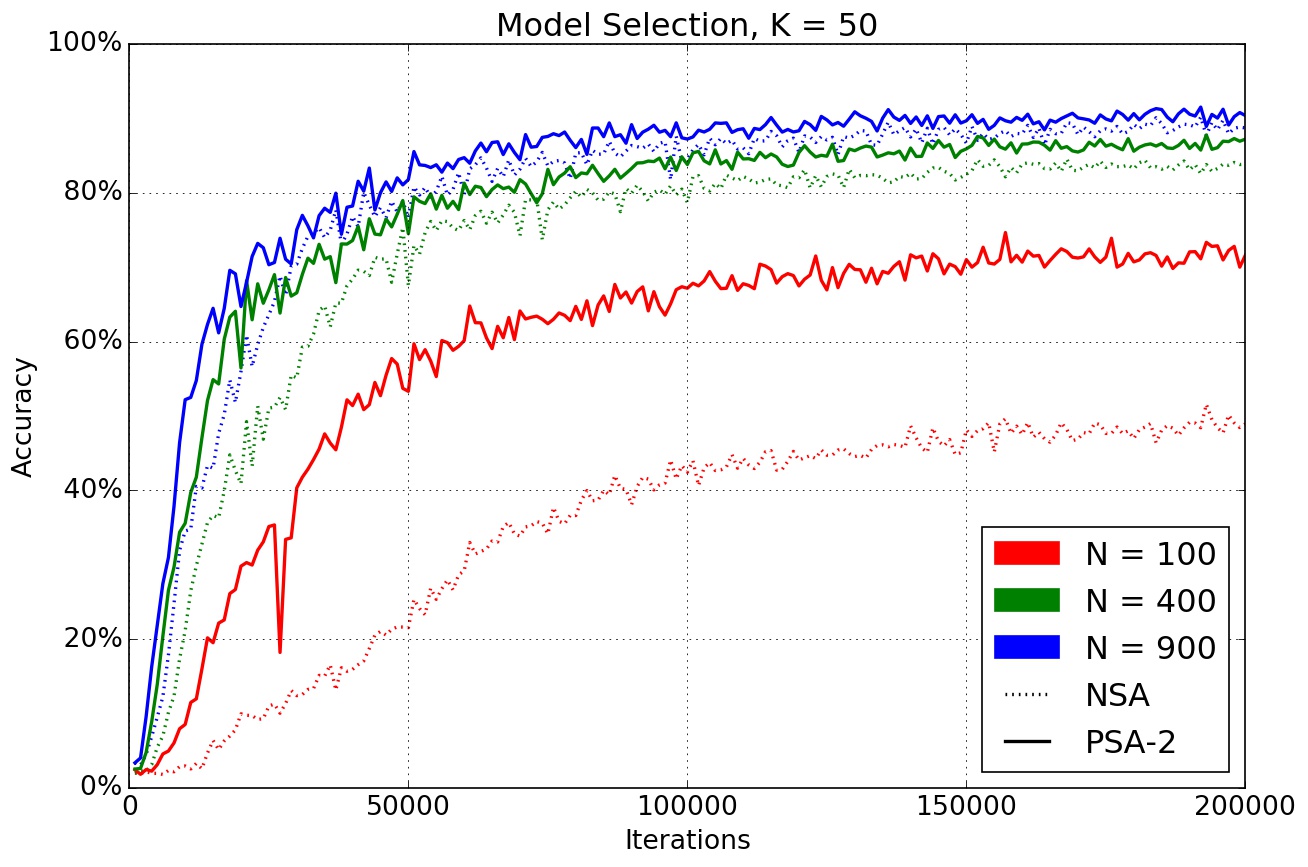} }}%
	\qquad
	\subfigure[]{{\includegraphics[width=6cm]{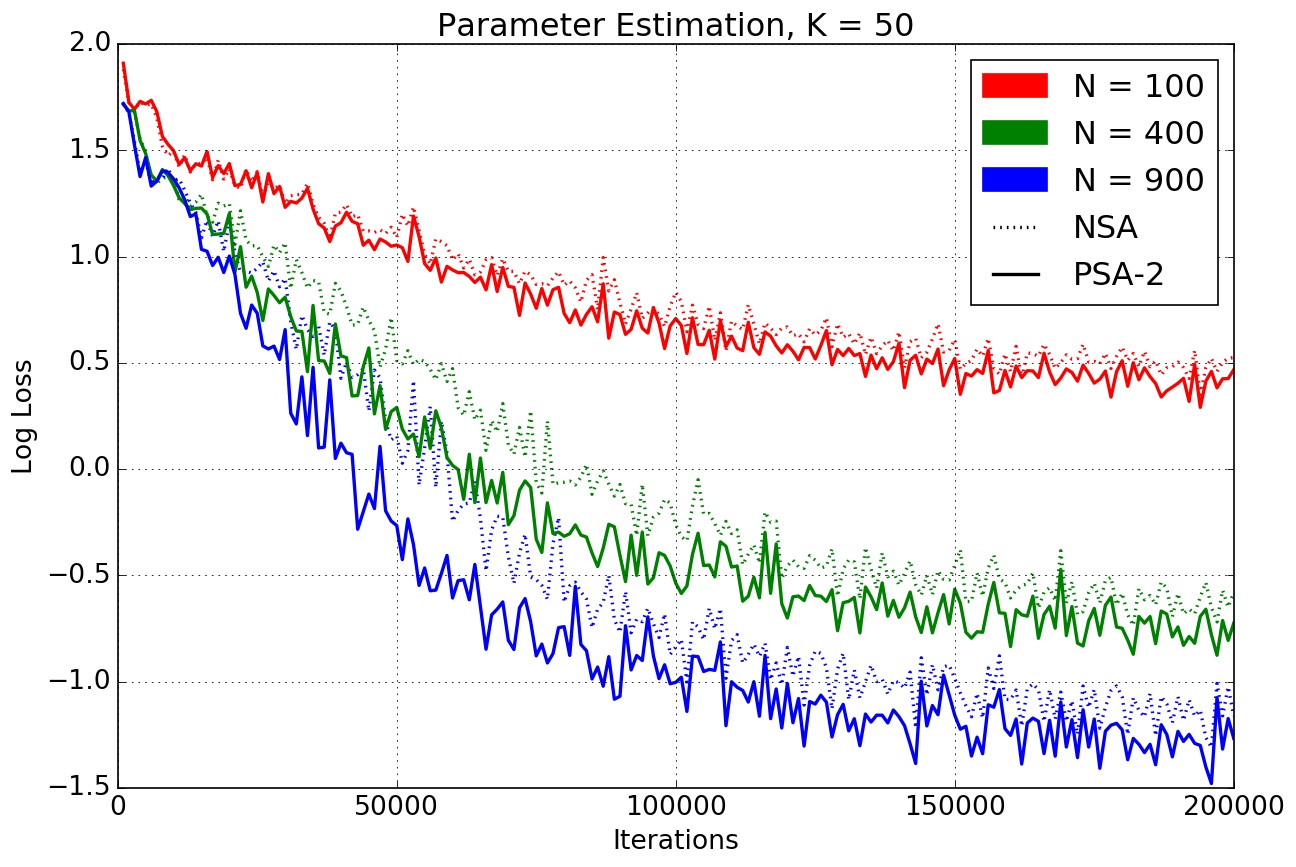} }}%
	\qquad
	\subfigure[]{{\includegraphics[width=6cm]{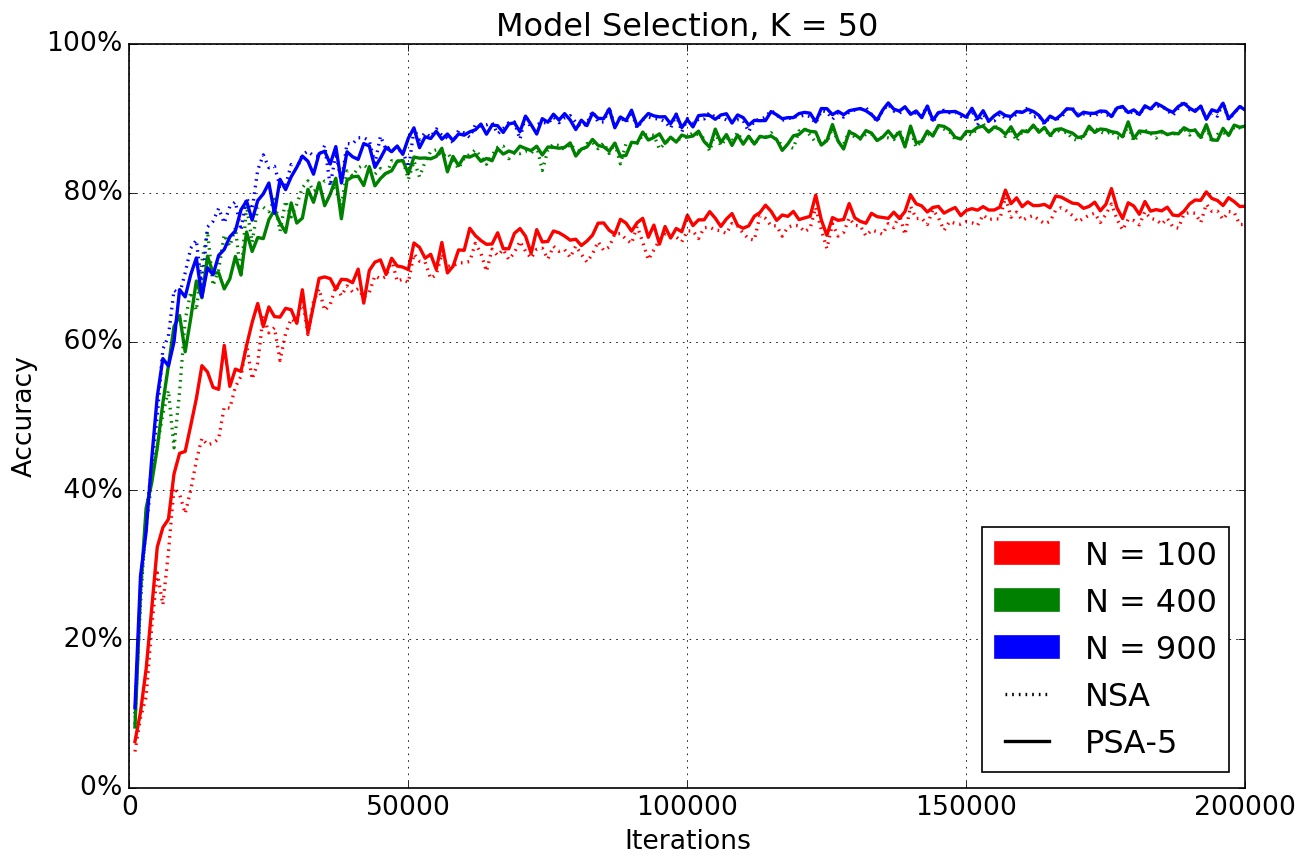} }}%
	\qquad
	\subfigure[]{{\includegraphics[width=6cm]{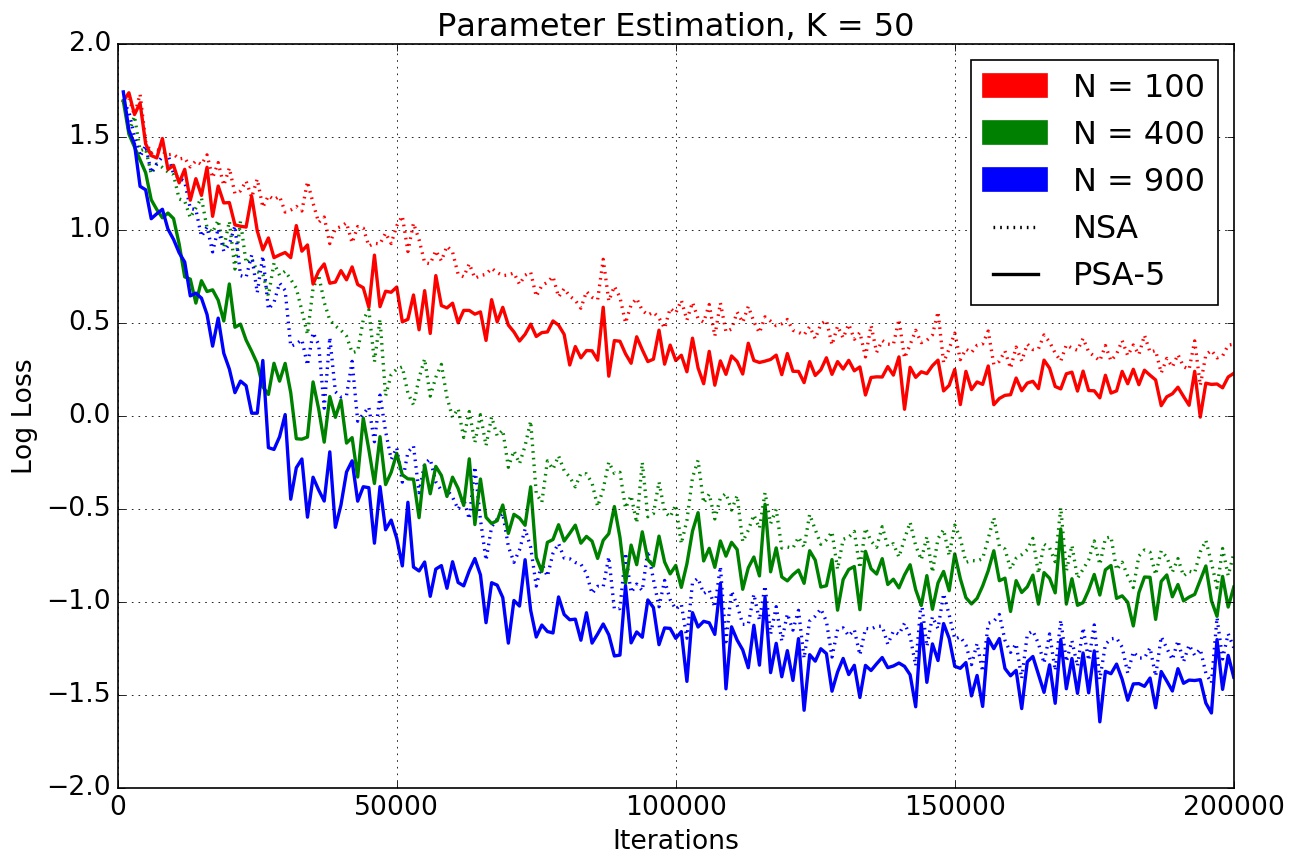} }}%
	\caption{Comparison between NSA and PSA-$l$ neural model selector and parameter estimator, different colours denote for different sample sizes, upper panel is for medium CNN architecture and lower panel is for large CNN architecture.}
	\label{fig:medium_50}
\end{figure}

{\em Impact of SA architectures on learning rate} Figure \ref{fig:medium_50} is used to compare the impacts of the NSA and PSA-$l$ SA architectures on the learning rates of the model selector and the parameter estimator, respectively. The medium and large CNN architectures are used, and all the three sample sizes are considered.  The upper panel is for medium CNN architecture while the lower panel is for large CNN architecture. The upper left panel of Figure \ref{fig:medium_50} plots the  accuracy of the model selector evaluated on the validation dataset against the number of iterations during the training process, whereas the upper right panel plots the log Huber loss of the parameter estimator. Solid curves are for the PSA-2 SA architecture, and dotted curves are for the NSA architectures. It is clear from the plots that the learning rate under PSA-2 is faster than that under NSA, indicating that sharing convolutional layers between the model selector and parameter estimator can expedite their training rates. Similar patterns could be found in other scenarios as showed in Figure S2, S3, S4 in Supplementary Document.

\begin{figure}[ht!]
	\centering
	\subfigure[]{{\includegraphics[width=6cm]{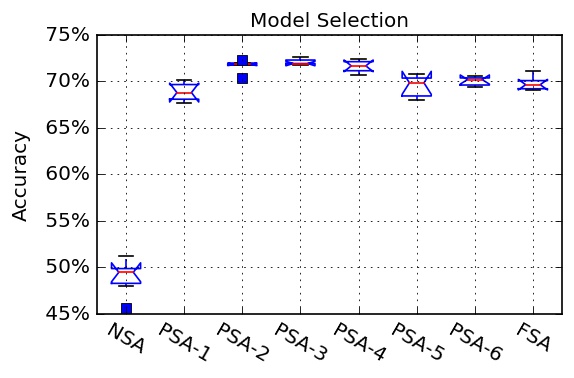} }}%
	\qquad
	\subfigure[]{{\includegraphics[width=6cm]{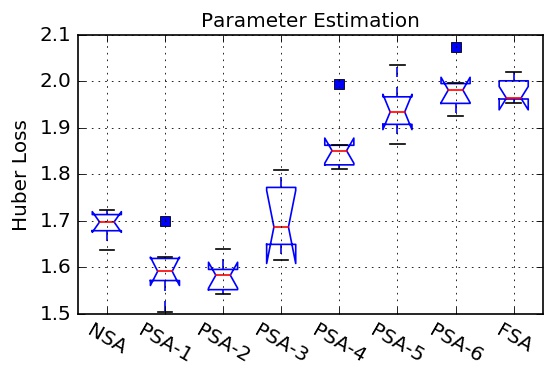} }}%
	\qquad
		\subfigure[]{{	\includegraphics[scale=0.3]{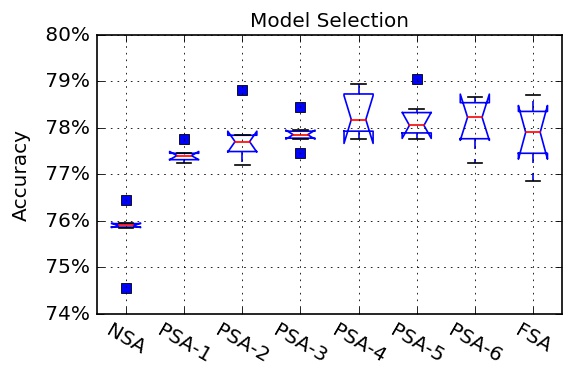} }}
		\subfigure[]{{	\includegraphics[scale=0.3]{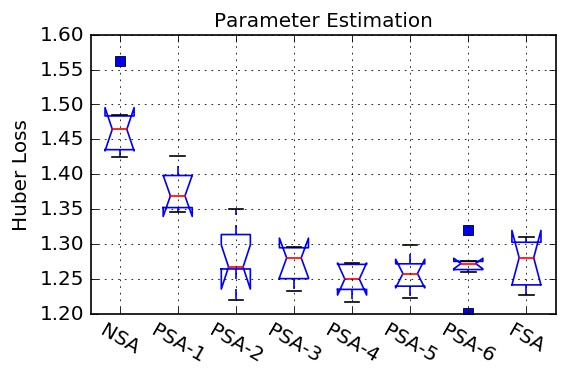} }}
	\caption{Information sharing  comparison for medium and large CNN architectures, $K=50$ and $N=100$. The upper panel is for medium CNN architecture and the lower panel is for large CNN architecture.}
	\label{fig:layerAnalysisMedium}
\end{figure}

\vspace{0.1in}
{\em How many layers should be shared?}  Figure \ref{fig:layerAnalysisMedium} shows the impact of the number of shared layers between the model selector and parameter estimator on their performances. We consider the scenario with $K=50$, $N=100$, and the medium and large CNN architectures, and vary the SA architectures from NSA to FSA. The left panel of Figure \ref{fig:layerAnalysisMedium} presents the boxplots of accuracy of the model selector under various SA architectures, whereas the right panel presents the boxplots of the Huber loss of the parameter estimator.  In terms of model selection accuracy, for medium CNN architecture, PSA-1 shows significant improvement over NSA, and PSA-2 further improves upon PSA-1, though the amount of improvement from PSA-1 to PSA-2 is small. PSA-3 performs almost the same as PSA-2, and further increasing the number of shared layers leads to slight decrease in selection accuracy.  In terms of estimation accuracy (i.e. Huber loss), we can observe similar patterns as the selection accuracy for NSA, PSA-1 and PSA-2. As the number of shared layers further increases, the estimation accuracy declines fairly fast. The results suggest that the PSA-2 SA architecture is optimal for both of the model selector and parameter estimator for the medium CNN architecture.  For the large CNN architecture, the optimal SA architecture turns out to be PSA-5 instead.

\subsection{ Comparison with conventional methods}\label{subsec: comparison}

We apply the three conventional model selection methods, the KS distance, BIC, and Bayes factor 
to the test datasets under the scenario with $K=20$, and compare their performances with that of the trained neural  model selector.   \ref{tabel:traditional} reports the accuracy of the three statistical methods as well as the trained neural model selector under various sample sizes. From the table, it is clear that the neural model selector outperforms the three statistical methods by a significant margin. 

In terms of accuracy in parameter estimation, conventional statistical estimators and the proposed neural estimator are not directly comparable. The former is based on the knowledge of the model, whereas the latter does not assume the underlying model is known.  If the model is known, then statistical methods such as the maximum likelihood estimation (MLE) method  are shown to enjoy certain optimality. For example, MLEs are asymptotically most efficient under some regularity conditions. If the model is unknown, then most conventional statistical methods are not applicable, but the neural parameter estimator can still work well.

\begin{table}[ht!]
	\caption{Comparison of model selection methods on model set with $K=20$.}	
	\begin{center}
		\begin{tabular}{ccccccc}
			\toprule
			& \multicolumn{2}{c}{\small{$N=100$}} & \multicolumn{2}{c}{\small{$N=400$}} & \multicolumn{2}{c}{\small{$N=900$}} \\  \cmidrule{2-7}
			&  \small{Top-1} & \small{Top-2} & \small{Top-1} & \small{Top-2} & \small{Top-1} & \small{Top-2} \\
			\midrule
			\small {KS distance} & 72.5\% & 83.2\% &  83.3\% &  85.0\%  & 84.7\% & 85.0\% \\ 
			\small {BIC}   & 69.9\% & 74.6\% &  74.7\% & 75.0\%  & 75.0\% & 75.0\% \\ 
			\small {Bayes factor} &  75.5\% & 84.8\% & 77.8\% & 83.3\%  & 70.0\% & 75.0\% \\ 
			\small {Neural selector} & 92.1\% & 99.2\% &  96.4\% &  99.7\%  & 97.9\% & 99.7\% \\ 
			\bottomrule
		\end{tabular}
	\end{center}
	\label{tabel:traditional}
\end{table}

\section {Neural selector for models with covariates}

In the  Introduction, we propose to use AI powered by deep neural networks to automate the SA process. In the previous sections, we develop the neural model selector and parameter estimator targeting only univariate models with single parameter. Our idea and proposed framework can be extended to handle more sophisticated models. In this section, we extend the neural selector to a group of commonly used simple regression models.

Let the model set $\mathcal{M}$ include the following seven regression models: simple linear regression model, Poisson regressoin model,  Logistic regression model,  Negative Binomial regression model, Lognormal regression model, Loglinear regression model, and multinormial regression model. Let $\{ (y_j, x_j)\}_{1\le j \le N}$ be a sample generated from one of the seven model. As before, the neural model selector is a CNN-based classifier that maps the sample to its generating model, and we will use labeled data systematically generated from the seven models to train this neural model selector.

The labeled data are generated as follows. For each regression model, we place an evenly spaced grid over its parameter space. For each vector of the parameter values on the grid,  1000 samples with sample size $N$ are randomly drawn from the model. The generated data are further partitioned into 70\% for training, 20\% for validation, and 10\% for test. We use the medium CNN architecture,  employ the Caffe to train the model selector, and further test the performance of the trained selector on the test dataset. The results show that the trained model selector can achieve 87.86\% in accuracy when the sample size is 100, and can achieve 97.86\% in accuracy when the sample size is 400.

\section{Real data example}

\begin{figure}%
	\centering
	\subfigure[]{{\includegraphics[width=6cm]{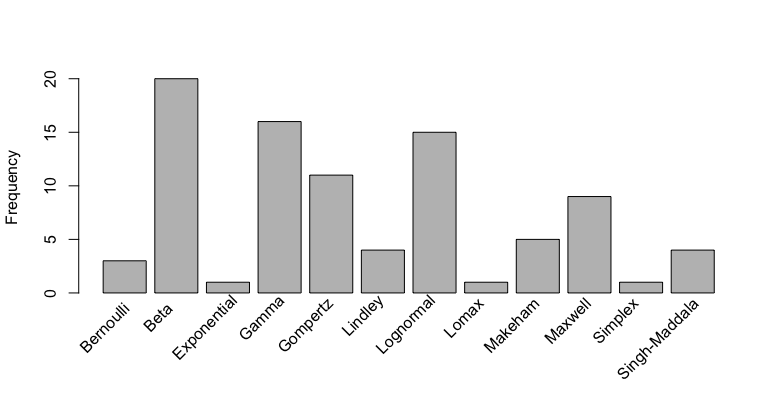} }}%
	\qquad
	\subfigure[]{{\includegraphics[width=6cm]{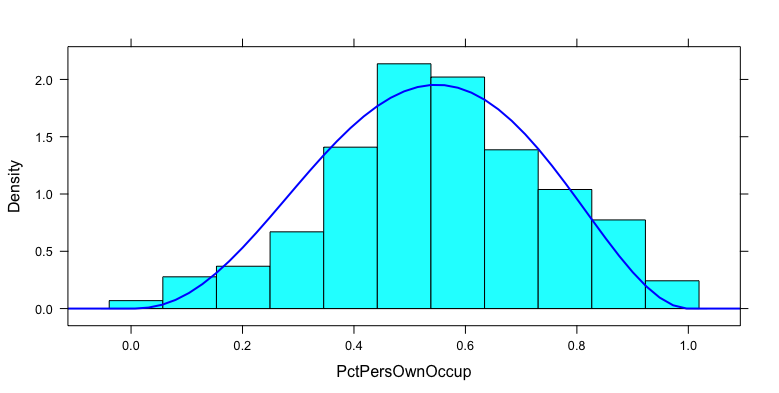} }}%
	\caption{Real data application}
	\label{fig:realdataanalysis}
\end{figure}

To demonstrate the applicability of our proposed methods, we use the neural model selector trained under the setting of $K=50$, 
$N=900$, large CNN, and PSA-5  to explore a real data set called the Communities and Crime Data Set from UC Irvine's machine learning repository (see \url{http://archive.ics.uci.edu/ml/datasets/Communities+and+Crime}). The data set originally contains 
1994 instances and 128 attributes. After we exclude the categorical variables and the variables with missing values, there are 90 real-valued predictor variables left.  For each of those 90 variables, we randomly draw a sub-sample of 900 observations,  apply the trained neural model selector to the sub-sample, and produce the result. We summarize the model selection results in the left panel of  Figure \ref{fig:realdataanalysis} for all 90 variables. We demonstrate the estimation result for the variable named PctPersOwnOccup (percent of people in owner occupied households) in the right panel of Figure \ref{fig:realdataanalysis}. The blue line is the density function with the estimated parameter value, while the histogram is based on the observations.

\section{Conclusion and future work}

In this paper, we have proposed and further developed the neural model selector and parameter estimator to automate model selection and parameter estimation, which are two major tasks in the SA process. Simulation study shows that the neural selector and estimator can be properly trained with simulated labeled data, and further demonstrate excellent performance. We consider this work a demonstration of the validity of our grand proposal that is to use DNNs to automate the entire SA process. There remains a lot of work we need to do before the grand proposal can be finally materialized. 

First, we will extend the neural model selector and parameter estimator to models with multiple parameters as well as regression models involving a large number of explanatory variables. Second, we will investigate how CNNs or other DNNs can be used to automate other tasks such as hypotheses testing and diagnostics of the SA process in the near future. Our ultimate goal to develop AI systems or software that can conduct principled SA for big data analytics without the need of human interventions.

\bibliography{nips}

\newpage

\appendix

\renewcommand\thefigure{\thesection S\arabic{figure}}    
\setcounter{figure}{0}

\renewcommand\thetable{\thesection S\arabic{table}}    
\setcounter{table}{0}

\begin{table}[h!]
	\begin{center}
		\caption{List of 50 models used in the simulation study}
		\scalebox{0.8}{
			\begin{tabular}{ cccc } 
				\toprule
				& Distribution  & Parameter of interest &  Other parameters \\
				\midrule
				1 & Bernoulli($p$)  & $p$ & \\
				2 & Discrete Uniform($N$)   & $N$ &  \\
				3 & Geometric($p$)  & $p$ & \\
				4 & Negtive Binomial($r$, $p$) & $r$ & $p=0.3$\\
				5 & Exponential($\lambda$) & $\lambda$ & \\
				6 & Normal($\mu$, $\sigma$) & $\mu$ & $\sigma=1$ \\
				7 & Poisson($\lambda$) & $\lambda$ & \\
				8 & Beta($\alpha$, $\beta$) & $\alpha$ & $\beta=3$ \\
				9 & Weibull($\lambda$, $k$) & $k$ & $\lambda=3$ \\
				10 & Double Exponential($\mu$, $\lambda$) & $\mu$  & $\lambda=3$ \\
				11 & Chi Square($k$) & $k$ & \\
				12 & F($d_1$, $d_2$) & $d_2$ & $d_1=3$\\
				13 & Gamma($\alpha$, $\beta$) & $\beta$ & $\alpha=0.5$ \\
				14 & Logistic($\mu$, $s$) & $\mu$ & $s=0.5$\\
				15 & Lognormal($\mu$, $\sigma$) & $\mu$ & $\sigma=0.5$\\
				16 & Pareto($x_m$, $\alpha$) & $x_m$ & $\alpha=2$ \\
				17 & Student's t($\nu$, $ncp$) & $\nu$ & $ncp=2$ \\
				18 & Uniform($a$, $a+2$) & $a$ & \\
				19 & Hypergeometric($m$, $n$, $k$) & $n$ & $m=3, k=2$ \\ 
				20 & Binomial($n$, $p$) & $n$ & $p=0.5$\\
				21 & One-Inflated Logarithmic($shape$, $pstr_1$) &   $shape$    & $pstr_1 = 0$  \\
				22 & Triangle($\theta$, $lower$, $upper$) & $\theta$       &  $lower = 0$, $upper = 33$ \\
				23 & Wilcoxon Signed Rank Statistic($n$) &  $n$      &  \\
				24 & Benini($y_0$, $shape$) &   $shape$     &  $y_0 = 1$ \\
				25 & Beta-Geometric($shape_1$, $shape_2$) &  $shape_2$      & $shape_1 = 5$  \\
				26 & Beta-Normal($shape_1$, $shape_2$, $mean$, $sd$) & $shape_1$       & $shape_2=10$, $mean=5$, $sd=11$  \\
				27 & Birnbaum-Saunders($scale$, $shape$) &  $shape$      & $scale = 1$  \\
				28 & Dagum($scale$, $shape_{1.a}$, $shape_{2.p}$) &    $shape_{1.a}$    & $scale = 1$, $shape_{2.p} = 2$  \\
				29 & Frechet($location$, $scale$, $shape$) &  $shape$      & $location = 0$, $scale = 1$  \\
				30 & Dirichlet($shape_1$, $shape_2$, $shape_3$) &  $shape_1$      & $shape_2 = 2$, $shape_3 = 4$ \\
				31 & Huber's Least Favourable($k$, $\mu$, $\sigma$) &  $k$      & $\mu = 0$, $\sigma = 1$ \\
				32 & Gumbel($location$, $scale$) &  $scale$      & $location = 1$ \\
				33 & Gompertz($scale$, $shape$) &  $shape$      & $scale = 1$  \\
				34 & Kumaraswamy($shape_1$, $shape_2$) &  $shape_2$      &  $shape_1 = 10$ \\
				35 & Laplace($location$, $scale$) &  $scale$      & $location = 5$ \\
				36 & Log-Gamma($location$, $scale$, $shape$) &  $scale$      & $location = 0$, $shape = 2$ \\
				37 & Lindley($\theta$) &  $\theta$      &  \\
				38 & Lomax($scale$, $shape_{3.q}$) &  $shape_{3.q}$      & $scale$ \\
				39 & Makeham($scale$, $shape$, $\epsilon$) &    $shape$    & $scale = 0$, $\epsilon = 0$  \\
				40 & Maxwell($rate$) &  $rate$      &  \\
				41 & Nakagami($scale$, $shape$, $Smallno$) &  $shape$      & $scale = 1$, $Smallno = 1.0e-6$  \\
				42 & Perks($scale$, $shape$) & $shape$       & $scale = 1$  \\
				43 & Rayleigh($scale$) &  $scale$      &  \\
				44 & Rice($\sigma$, $vee$) & $vee$       & $\sigma = 1$ \\
				45 & Simplex($\mu$, $dispersion$) &  $dispersion$      & $\mu = 0.5$  \\
				46 & Singh-Maddala($scale$, $shape_{1.a}$, $shape_{3.q}$) &  $shape_3.q$      & $scale = 1$, $shape_{1.a} = 5$ \\
				47 & Skellam($\mu_1$, $\mu_2$) &    $\mu_2$    & $\mu_1 = 5$  \\
				48 & Tobit($mean$, $sd$, $lower$, $upper$) &  $mean$      & $sd = 1$, $lower = 0$, $upper = Inf$  \\
				49 & Paralogistic($scale$, $shape_{1.a}$) &  $scale_{1.a}$      & $scale = 1$  \\
				50 & Zipf($N$, $shape$) &  $shape$      & $N = 10$  \\
				\bottomrule
			\end{tabular}
		}
	\end{center}
	\label{DistributionList}
\end{table}

\begin{figure}[h!]
	\centering
	\includegraphics[width=15cm, height=15cm]{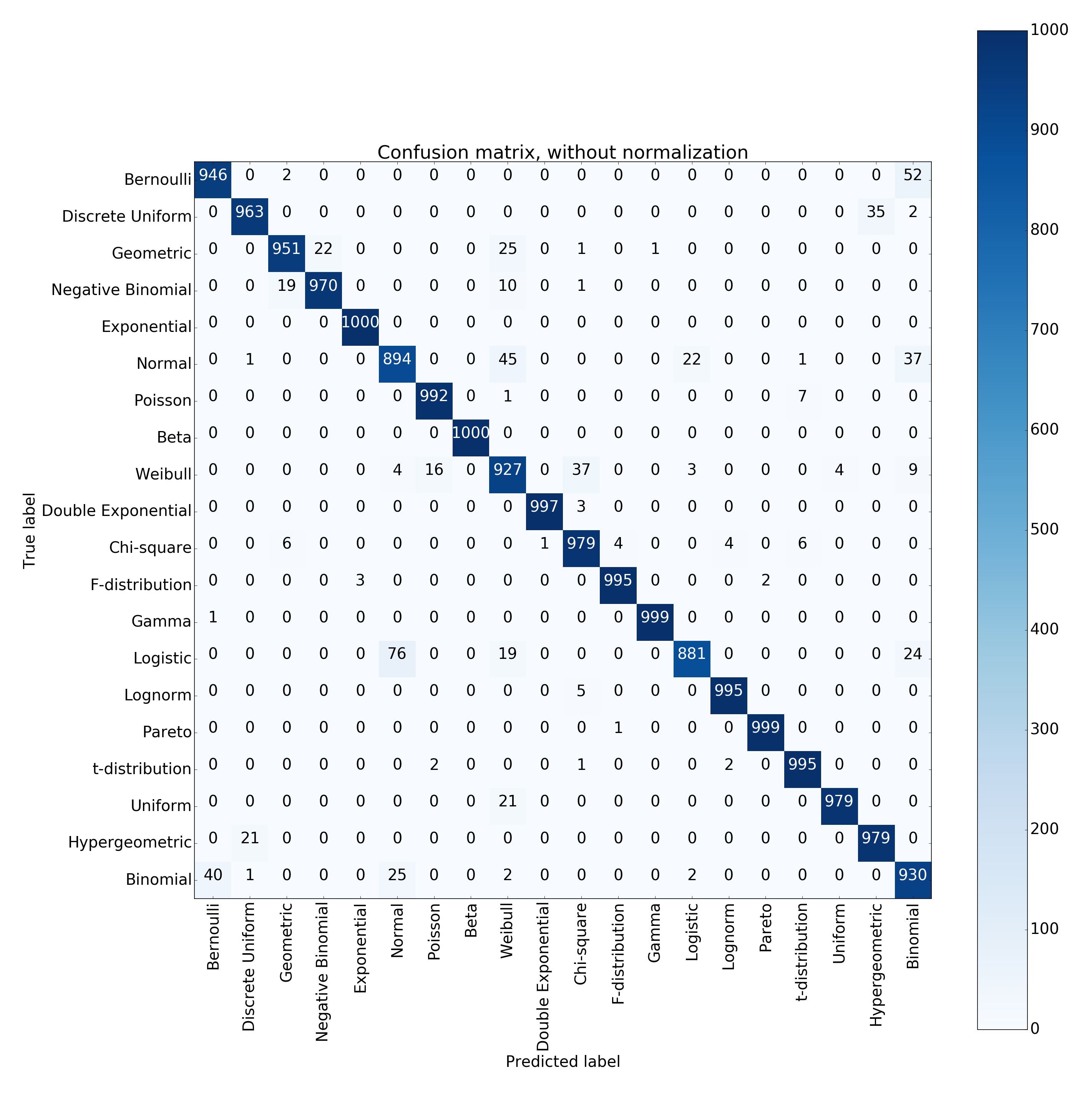}
	\caption{Confusion matrix based on large CNN and PSA-5 neural model selector on test dataset with  $K=20$}
	\label{confusionmatrix}
\end{figure}

\begin{figure}[]
	\centering
	\subfigure[]{{\includegraphics[width=6cm, height=6cm]{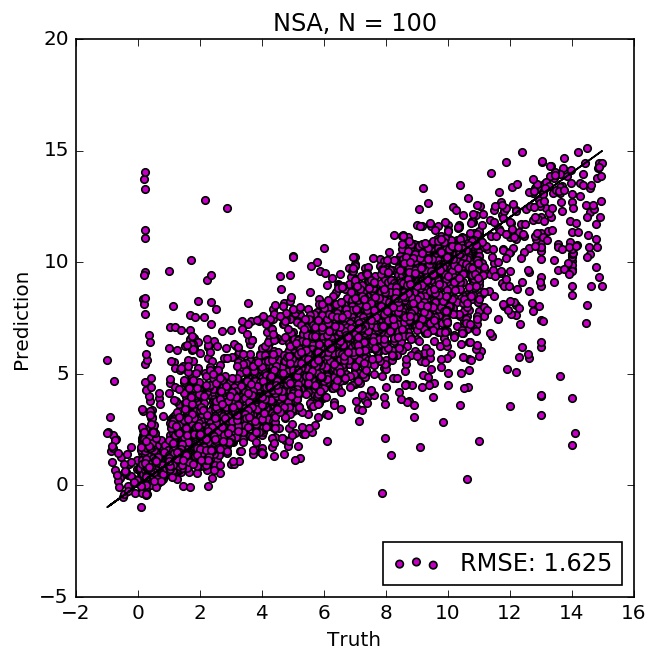} }}
	\qquad
	\subfigure[]{{ \includegraphics[width=6cm, height=6cm]{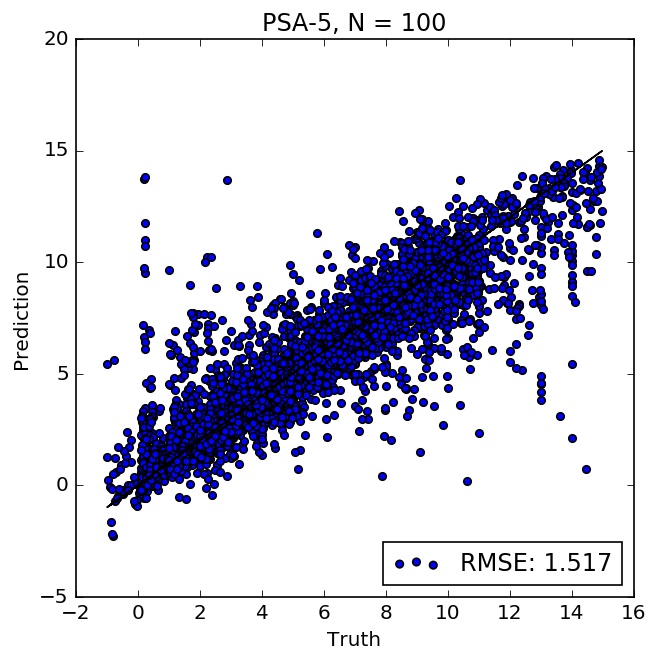} }}
	
	\subfigure[]{{	\includegraphics[width=6cm, height=6cm]{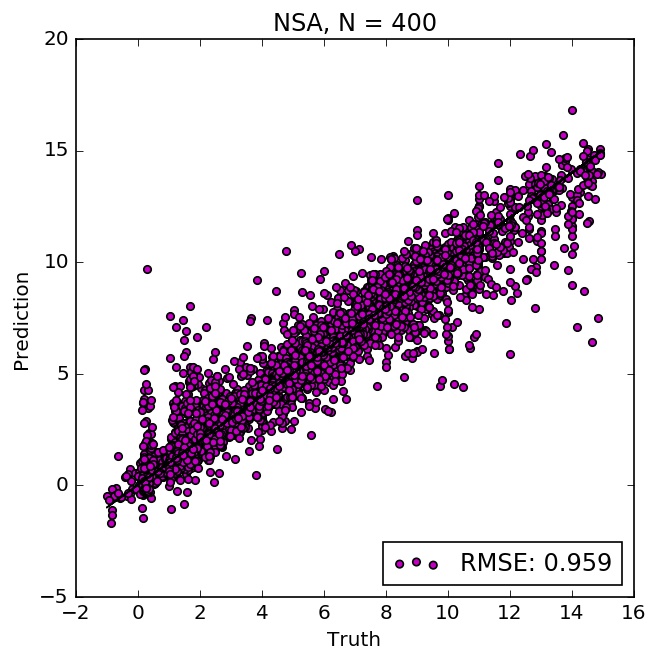} }}
	\qquad
	\subfigure[]{{	\includegraphics[width=6cm, height=6cm]{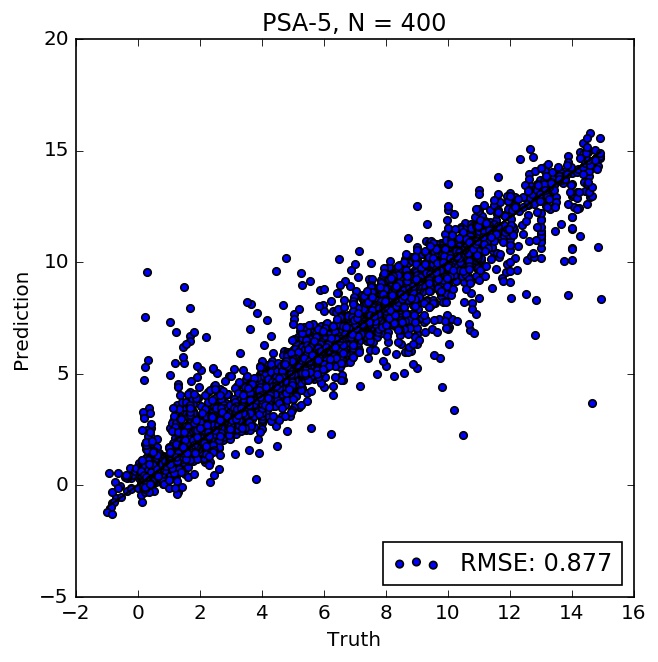} }}
	
	\subfigure[]{{	\includegraphics[width=6cm, height=6cm]{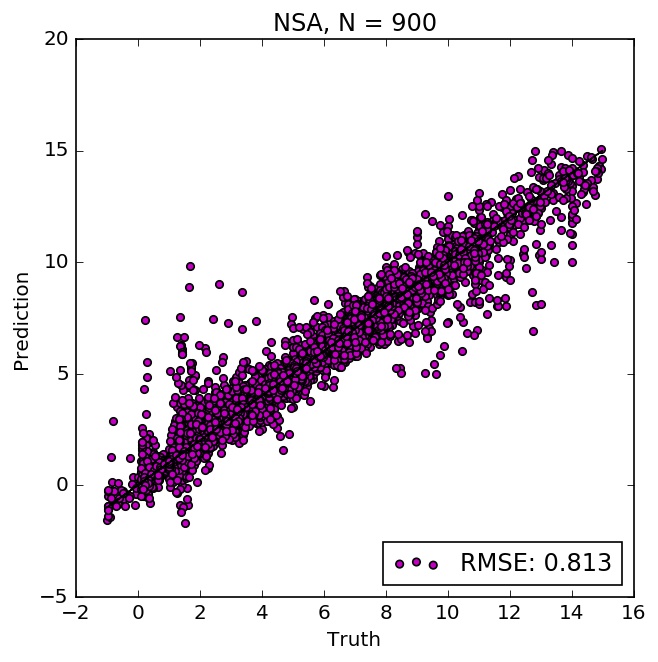} }}
	\qquad
	\subfigure[]{{	\includegraphics[width=6cm, height=6cm]{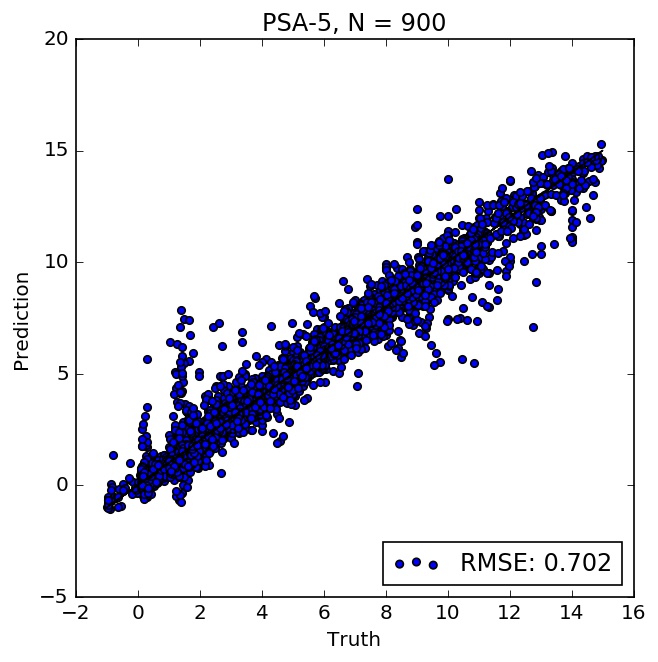}  }}
	
	\caption{Parameter estimation results on the  test dataset with $K=50$. The left panel is the results estimated by large CNN and NSA parameter estimator, while the right panel is estimated by large CNN and PSA-5 parameter estimator. The x-axis in each plot is the ground truth for the parameters and the y-axis is the estimation.}
	\label{scatter1}
\end{figure}

\begin{figure}
	\centering
	\subfigure[Model Selection]{{	\includegraphics[width=7cm, height=14cm]{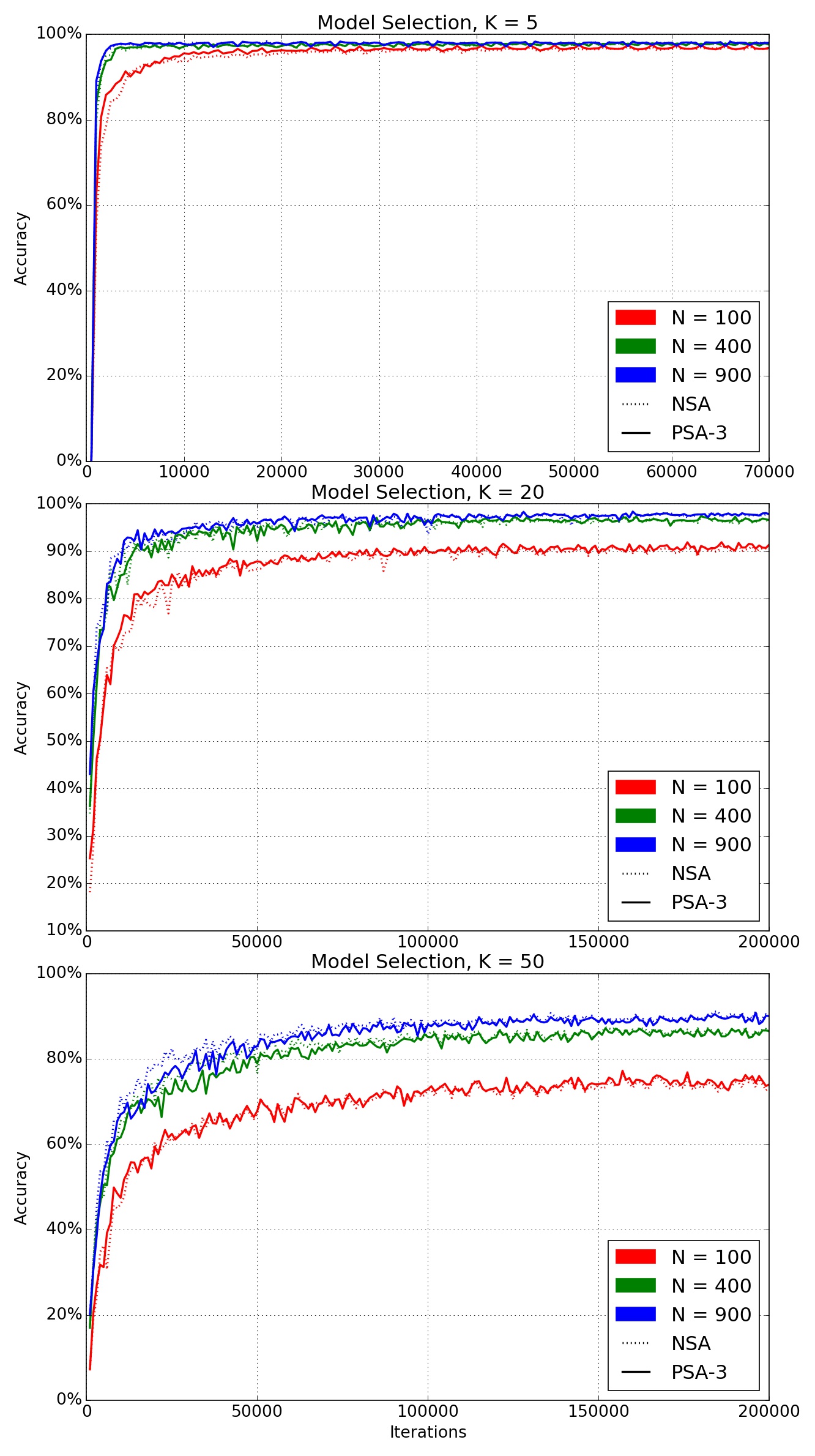} 	\label{fig:sub1} }}
	\subfigure[Parameter Estimation]{{	\includegraphics[width=7cm, height=14cm]{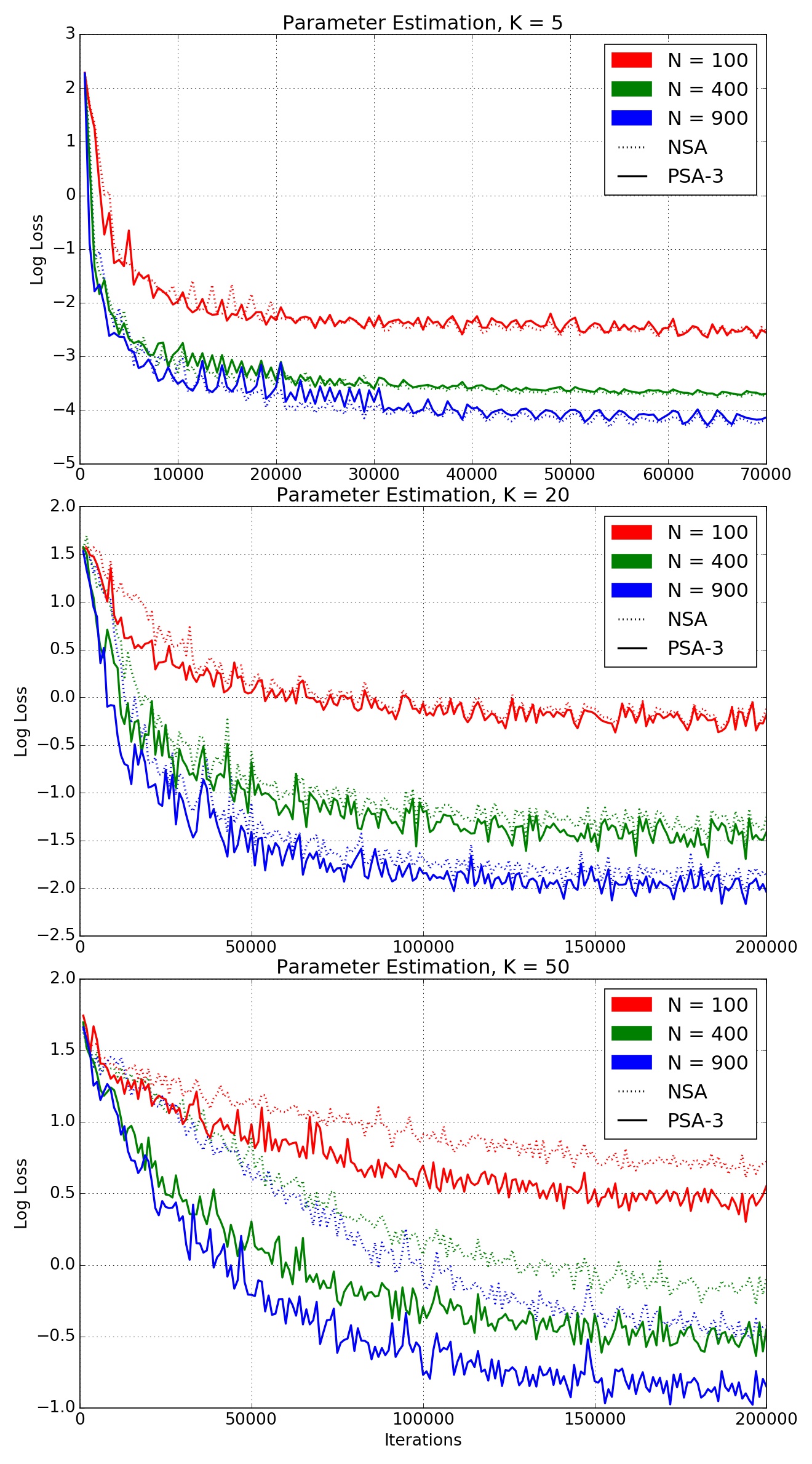}	\label{fig:sub2}  }}
	
	\caption{Comparison between NSA and PSA-3:  learning curves of selection accuracy and estimation Huber loss for different samples sizes and different number of candidate models. This is based on small CNN architecture. The left panel plots the selection accuracy of the model selector evaluated on the validation dataset against the number of iterations during the training process, whereas the right panel plots the log Huber loss of the parameter estimator on the validation dataset against the number of iterations during the training process. Solid curves are for the PSA-3, and dotted curves are for the NSA. Different colors denote for different sample sizes.
	}
	\label{app:compareNSAandPSAsmall}
\end{figure}

\begin{figure}
	\centering
	\subfigure[Model Selection]{{	\includegraphics[width=7cm, height=14cm]{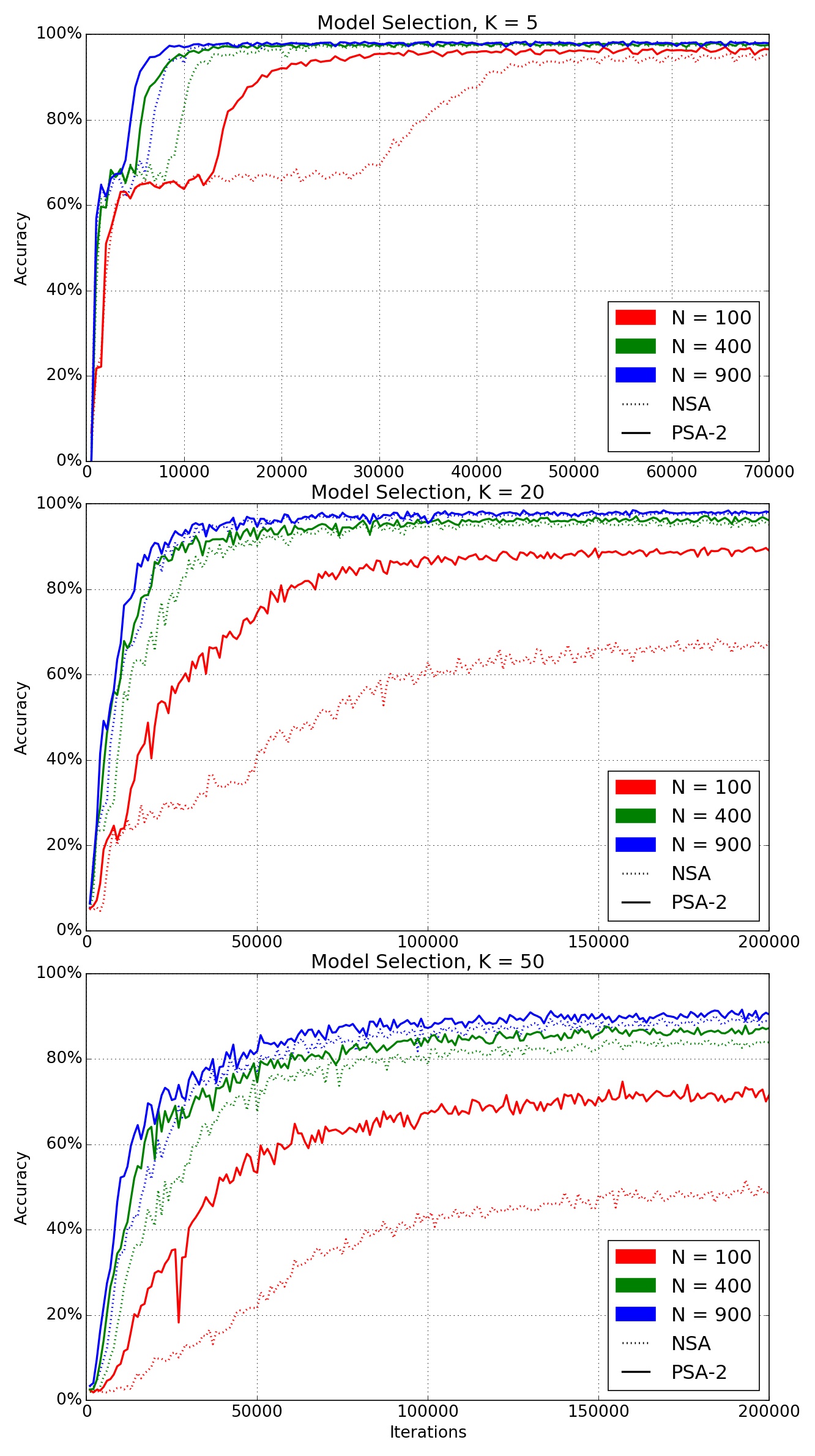} \label{fig:sub1} }}
	\subfigure[Parameter Estimation]{{	\includegraphics[width=7cm, height=14cm]{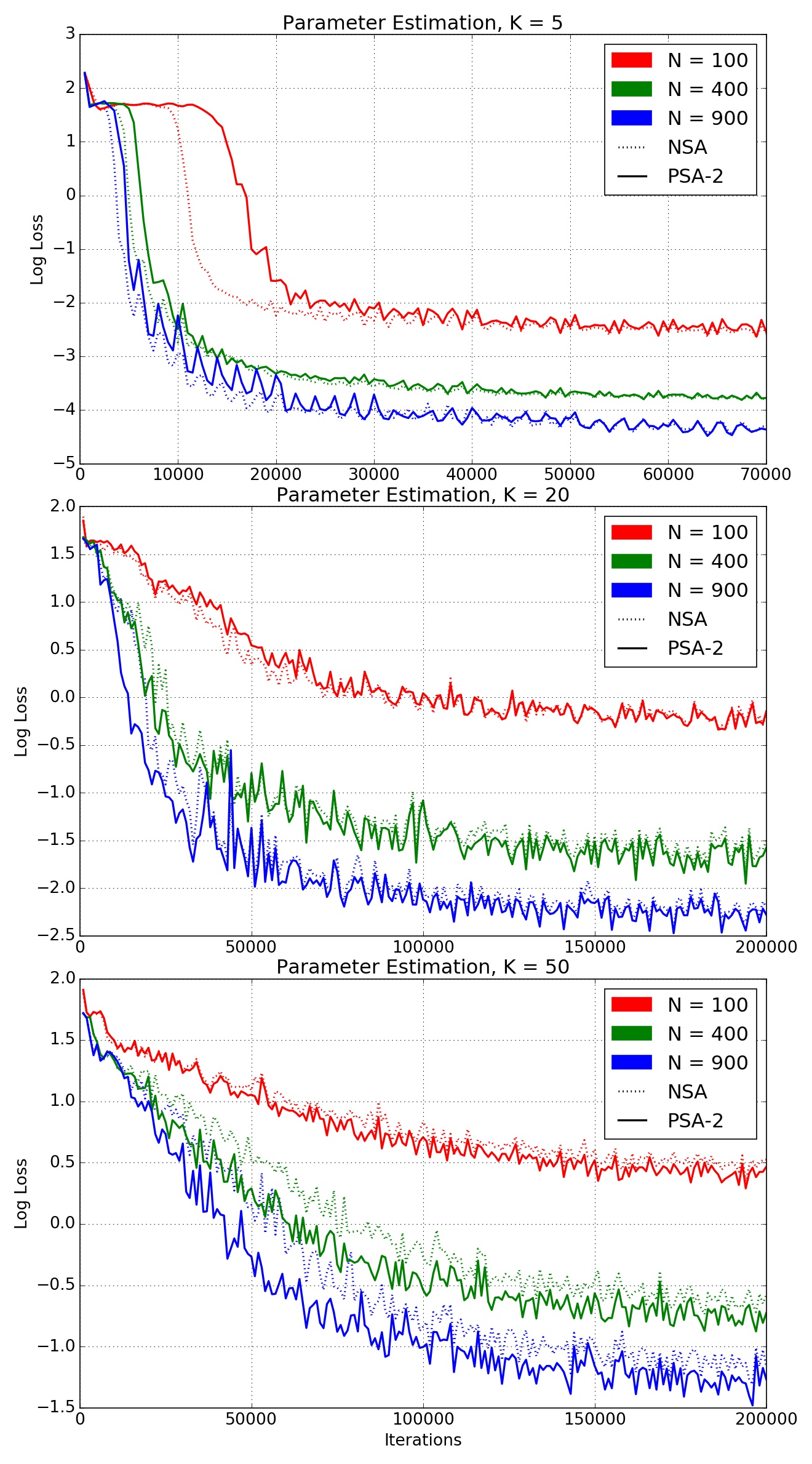} \label{fig:sub2} }}
	
	\caption{ Comparison between NSA and PSA-2:  learning curves of selection accuracy and estimation Huber loss for different samples sizes and different number of candidate models. This is based on medium CNN architecture. The left panel plots the selection accuracy of the model selector evaluated on the validation dataset against the number of iterations during the training process, whereas the right panel plots the log Huber loss of the parameter estimator on the validation dataset against the number of iterations during the training process. Solid curves are for the PSA-2, and dotted curves are for the NSA. Different colors denote for different sample sizes.
	}
	\label{app:compareNSAandPSAmedium}
\end{figure}

\begin{figure}[h]
	\centering
	\subfigure[Model Selection]{{	\includegraphics[width=7cm, height=14cm]{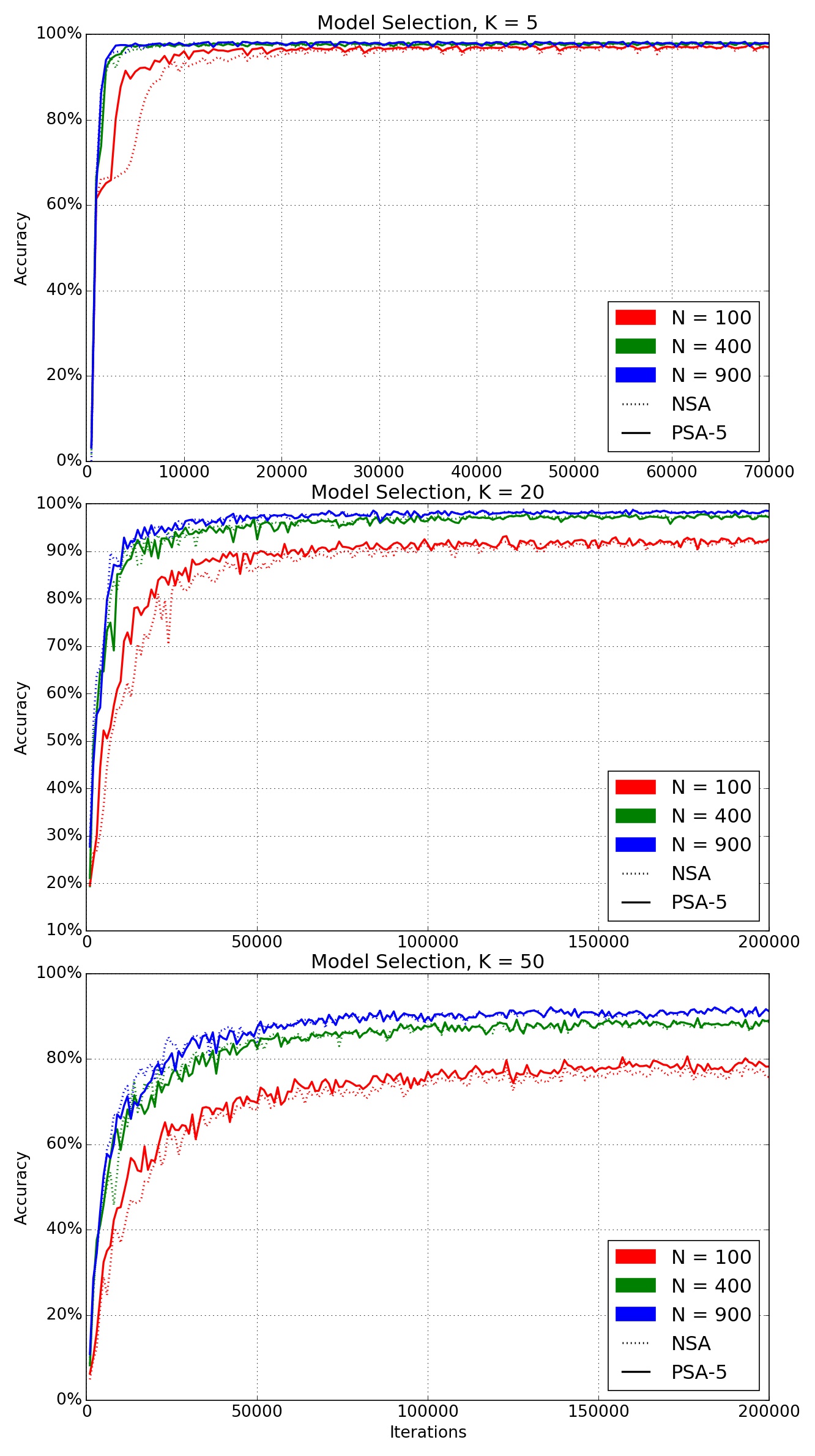}
			\label{fig:sub1} }}
	\subfigure[Parameter Estimation]{{	\includegraphics[width=7cm, height=14cm]{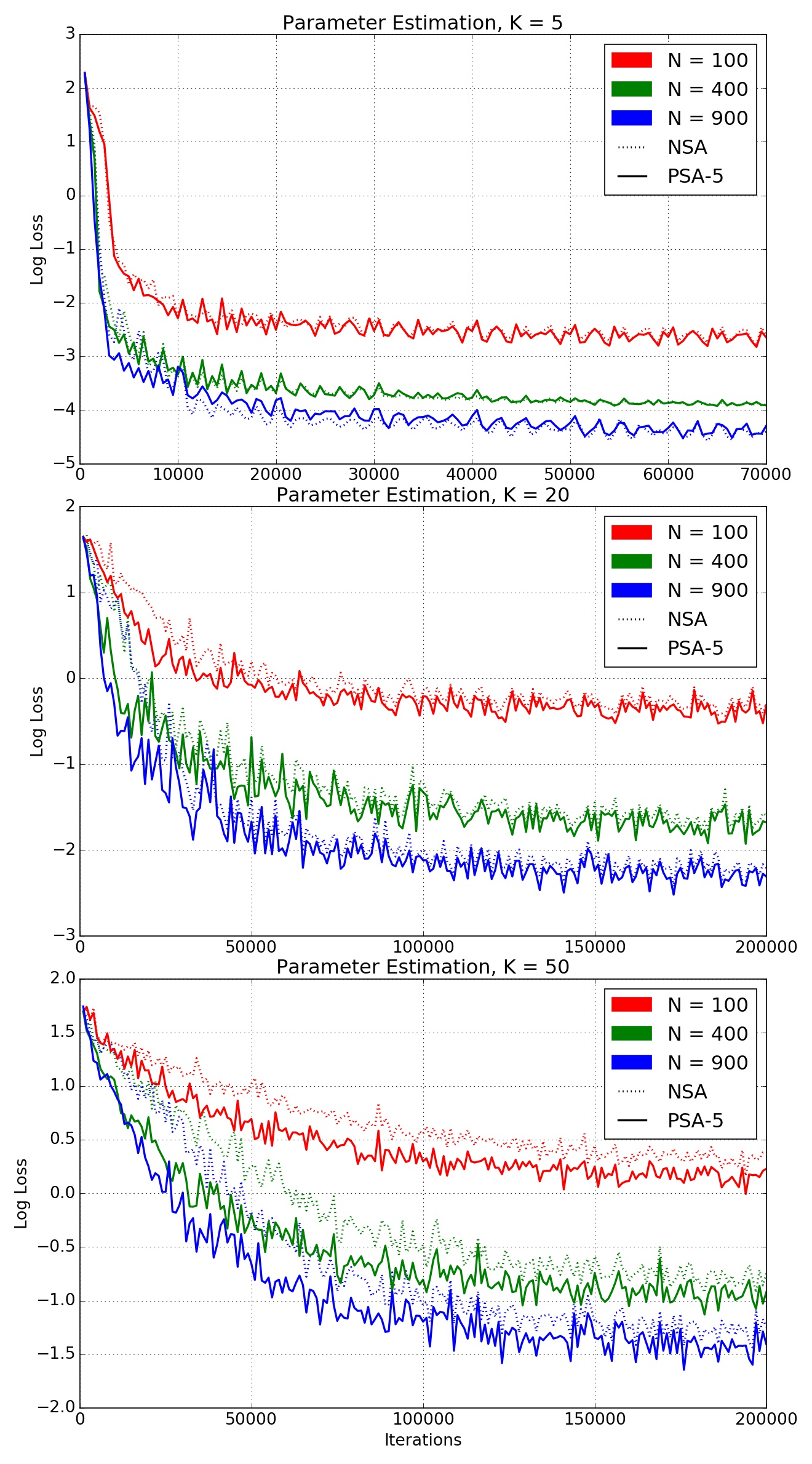}
			\label{fig:sub2} }}
	\caption{ 
		Comparison between NSA and PSA-5:  learning curves of selection accuracy and estimation Huber loss for different samples sizes and different number of candidate models. This is based on large CNN architecture. The left panel plots the selection accuracy of the model selector evaluated on the validation dataset against the number of iterations during the training process, whereas the right panel plots the log Huber loss of the parameter estimator on the validation dataset against the number of iterations during the training process. Solid curves are for the PSA-5, and dotted curves are for the NSA. Different colors denote for different sample sizes.
	}
	\label{app:compareNSAandPSAlarge}
\end{figure}

\begin{figure}
	\centering
	\subfigure[]{{	\includegraphics[width=4.5cm, height=4.5cm]{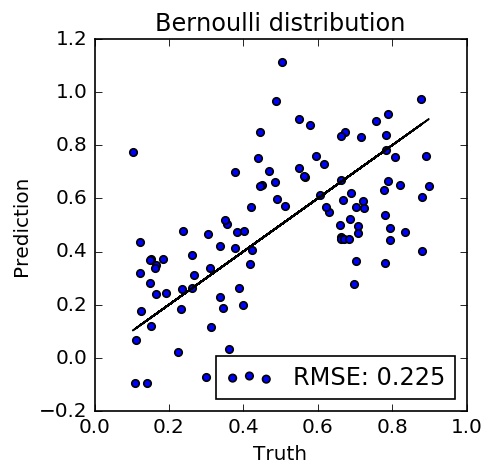}  }}
	\subfigure[]{{	\includegraphics[width=4.5cm, height=4.5cm]{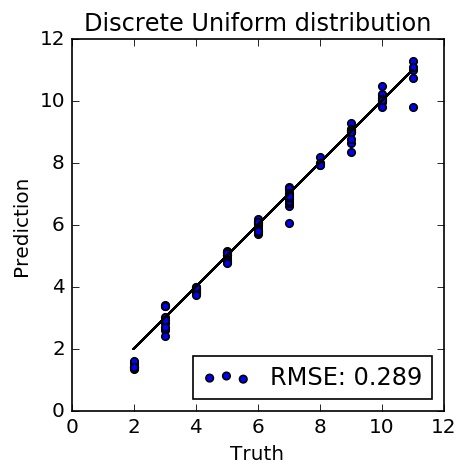}  }}
	\subfigure[]{{	\includegraphics[width=4.5cm, height=4.5cm]{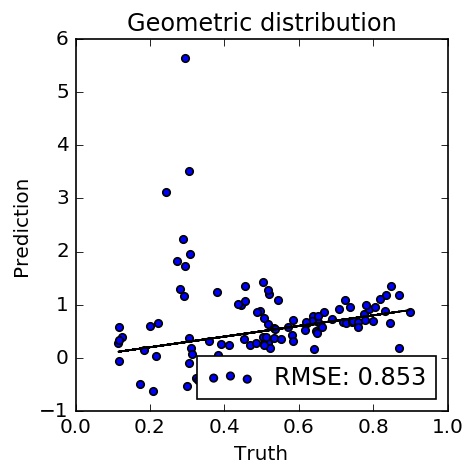} }}
	\subfigure[]{{	\includegraphics[width=4.5cm, height=4.5cm]{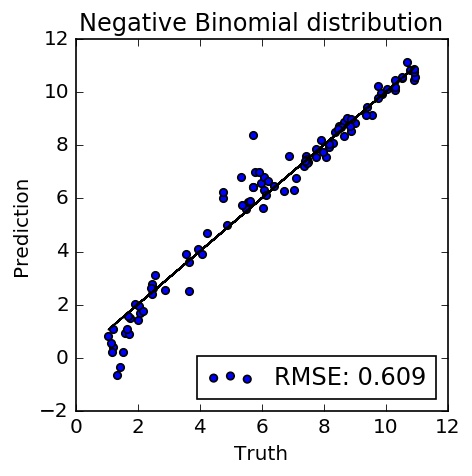} }}
	\subfigure[]{{	\includegraphics[width=4.5cm, height=4.5cm]{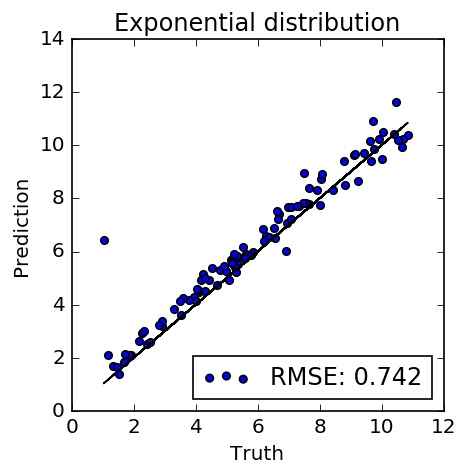} }}
	\subfigure[]{{	\includegraphics[width=4.5cm, height=4.5cm]{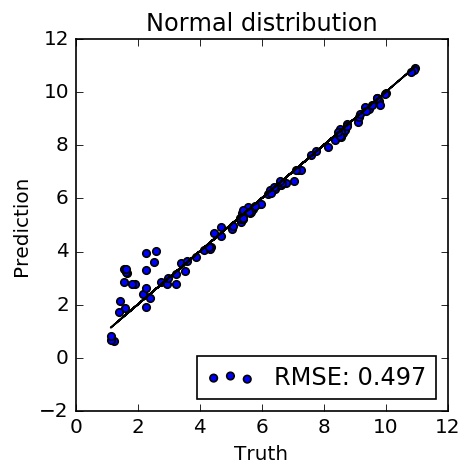} }}
	\subfigure[]{{	\includegraphics[width=4.5cm, height=4.5cm]{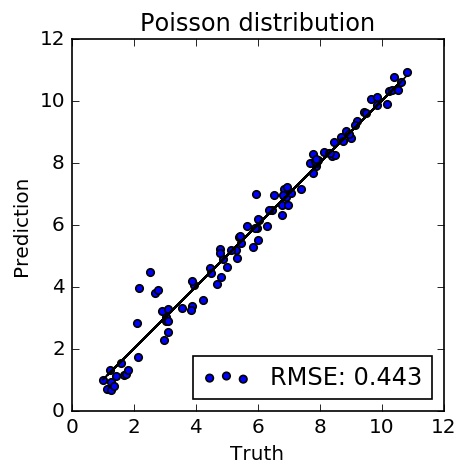} }}
	\subfigure[]{{	\includegraphics[width=4.5cm, height=4.5cm]{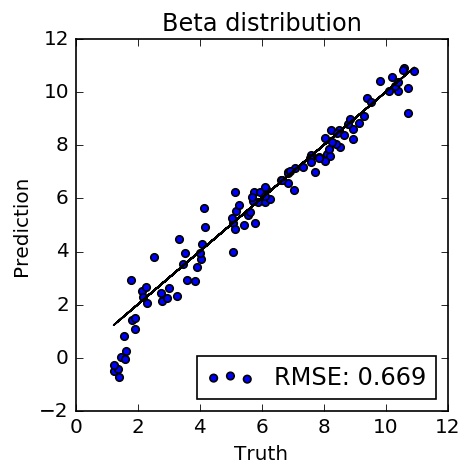} }}
	\subfigure[]{{	\includegraphics[width=4.5cm, height=4.5cm]{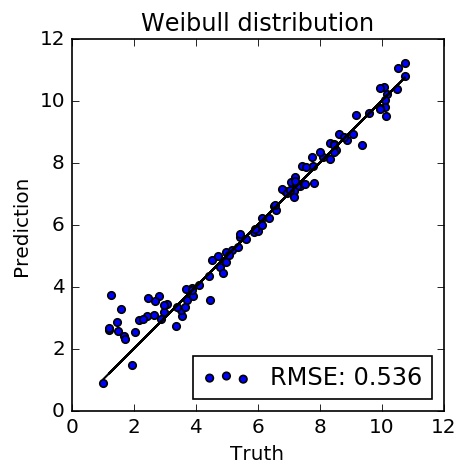} }}
	\subfigure[]{{	\includegraphics[width=4.5cm, height=4.5cm]{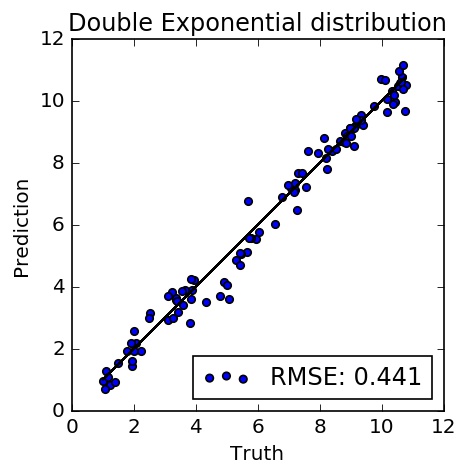} }}
	\subfigure[]{{\includegraphics[width=4.5cm, height=4.5cm]{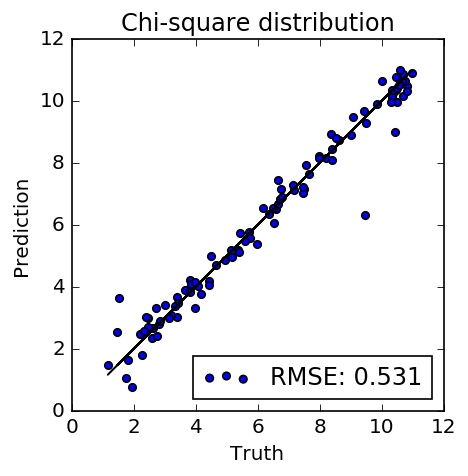} }}
	\subfigure[]{{	\includegraphics[width=4.5cm, height=4.5cm]{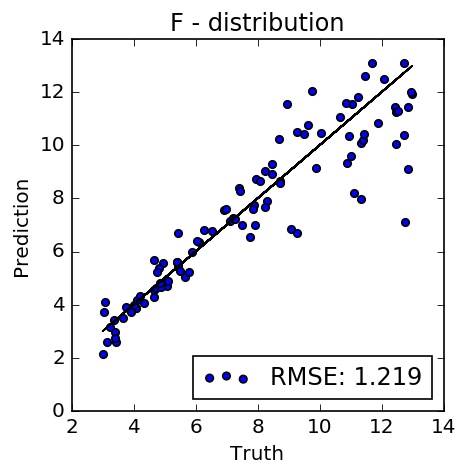} }}
	\caption{Distribution wise performance of PSA-5 neural parameter estimator under large CNN on the test dataset with $N=900$. RMSE is reported for each distribution. Part 1.}
	\label{scatterplotpart1}
\end{figure}

\begin{figure}
	\centering
	\subfigure[]{{	\includegraphics[width=4.5cm, height=4.5cm]{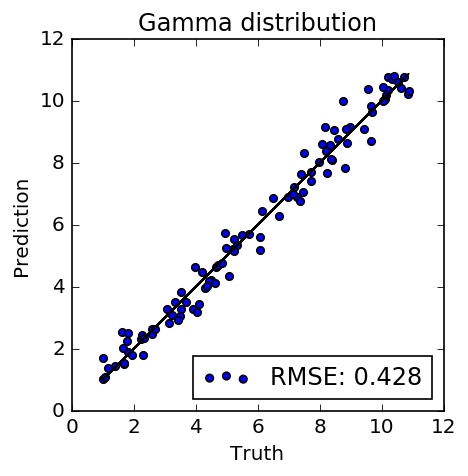} }}
	\subfigure[]{{	\includegraphics[width=4.5cm, height=4.5cm]{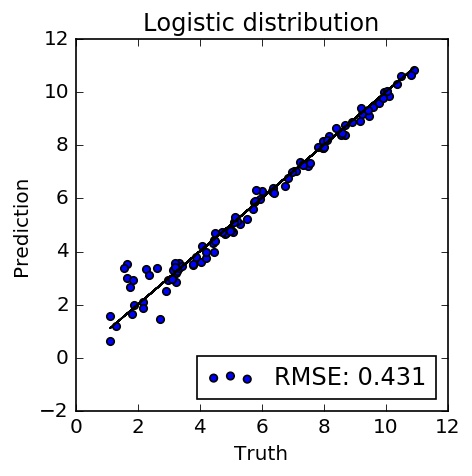} }}
	\subfigure[]{{	\includegraphics[width=4.5cm, height=4.5cm]{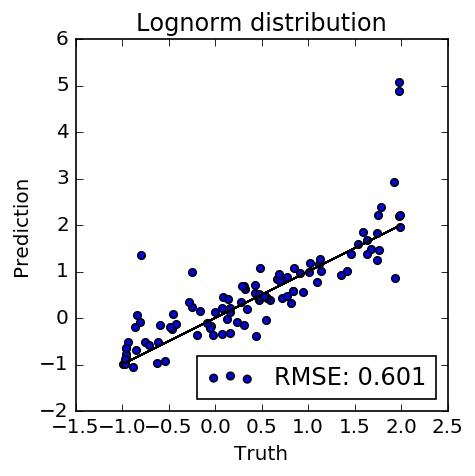} }}
	\subfigure[]{{	\includegraphics[width=4.5cm, height=4.5cm]{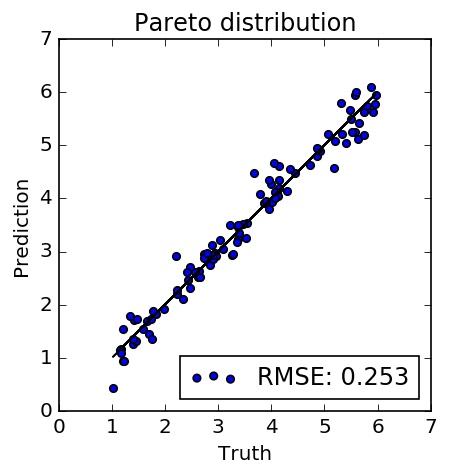} }}
	\subfigure[]{{	\includegraphics[width=4.5cm, height=4.5cm]{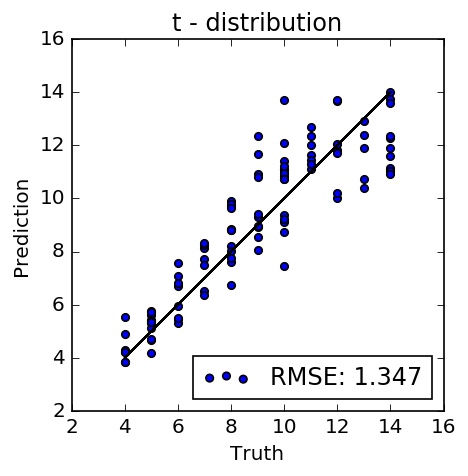} }}
	\subfigure[]{{	\includegraphics[width=4.5cm, height=4.5cm]{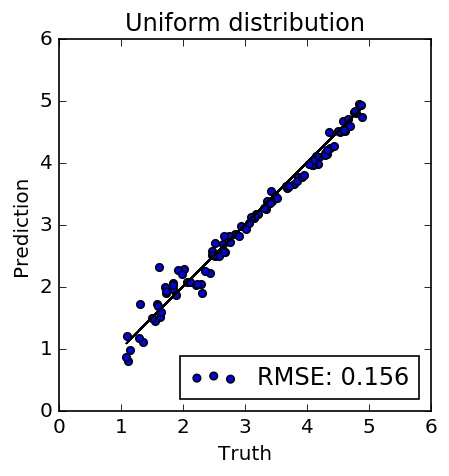} }}
	\subfigure[]{{	\includegraphics[width=4.5cm, height=4.5cm]{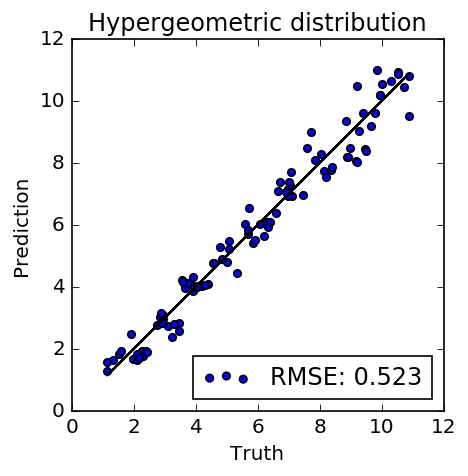} }}
	\subfigure[]{{	\includegraphics[width=4.5cm, height=4.5cm]{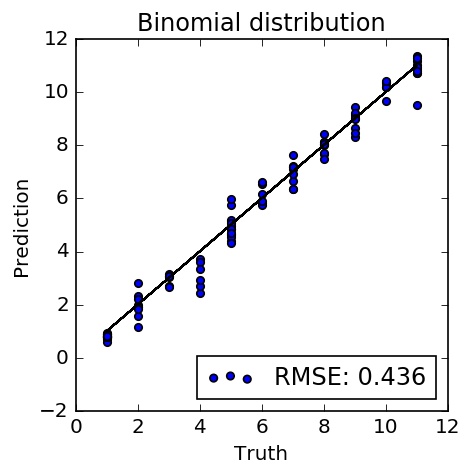} }}
	\subfigure[]{{	\includegraphics[width=4.5cm, height=4.5cm]{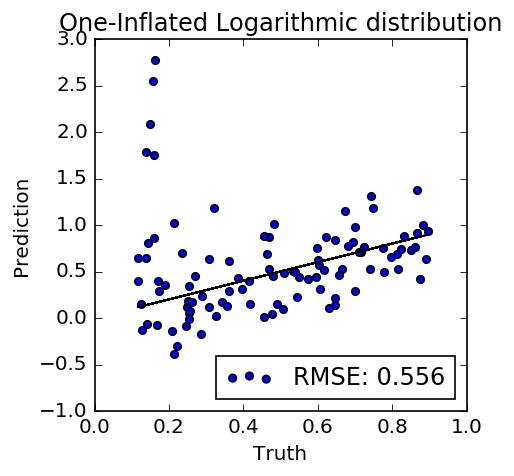} }}
	\subfigure[]{{	\includegraphics[width=4.5cm, height=4.5cm]{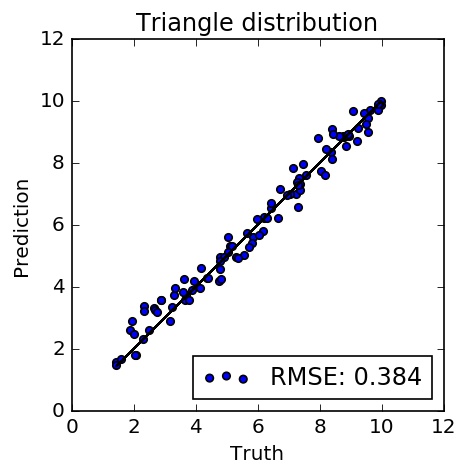} }}
	\subfigure[]{{	\includegraphics[width=4.5cm, height=4.5cm]{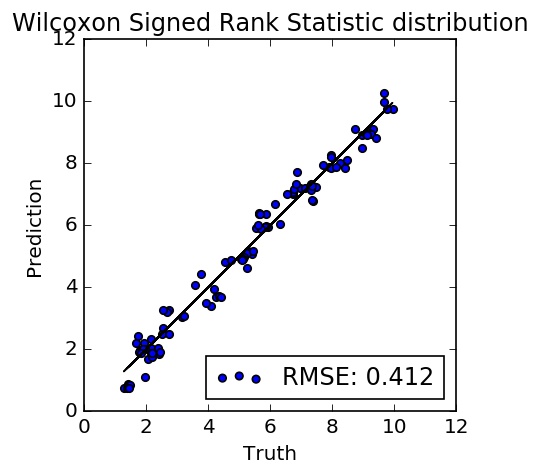} }}
	\subfigure[]{{	\includegraphics[width=4.5cm, height=4.5cm]{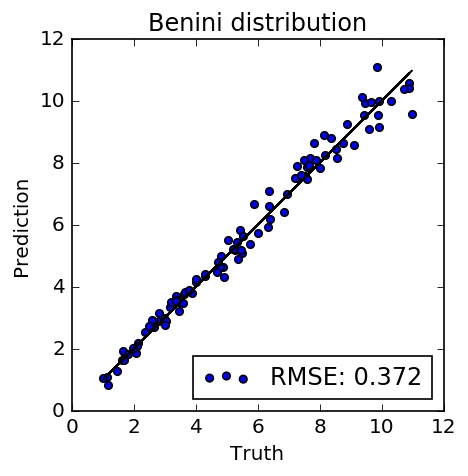} }}
	
	\caption{Distribution wise performance of PSA-5 neural parameter estimator under large CNN on the test dataset with $N=900$. RMSE is reported for each distribution. Part 2.}
	\label{scatterplotpart2}
\end{figure}

\begin{figure}
	\centering
	\subfigure[]{{		\includegraphics[width=4.5cm, height=4.5cm]{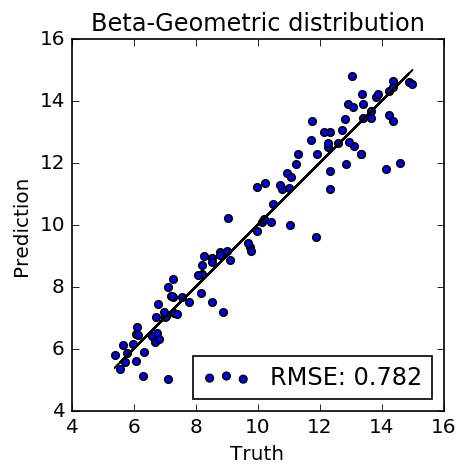} }}
	\subfigure[]{{		\includegraphics[width=4.5cm, height=4.5cm]{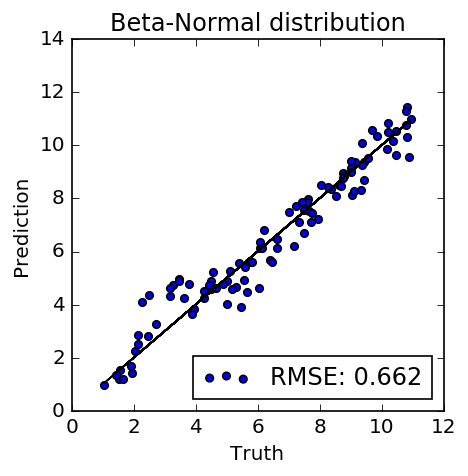} }}
	\subfigure[]{{		\includegraphics[width=4.5cm, height=4.5cm]{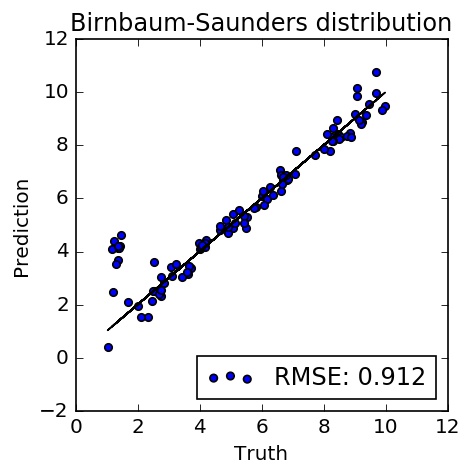} }}
	\subfigure[]{{		\includegraphics[width=4.5cm, height=4.5cm]{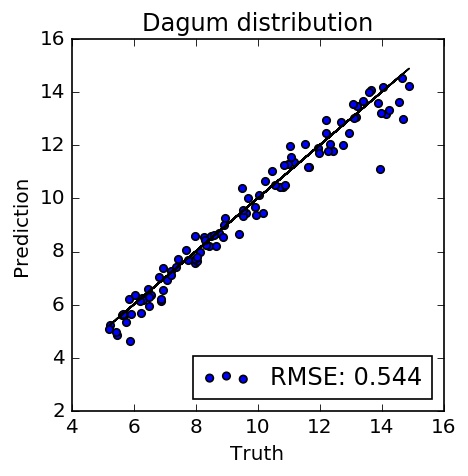} }}
	\subfigure[]{{		\includegraphics[width=4.5cm, height=4.5cm]{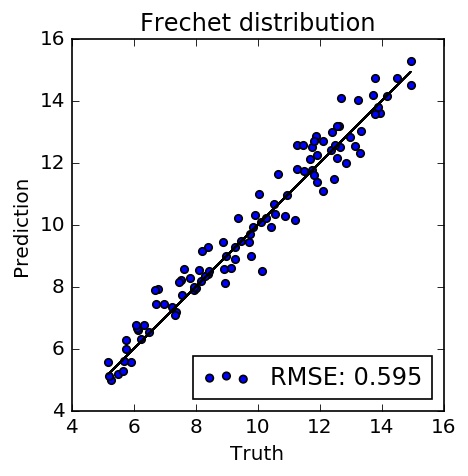} }}
	\subfigure[]{{		\includegraphics[width=4.5cm, height=4.5cm]{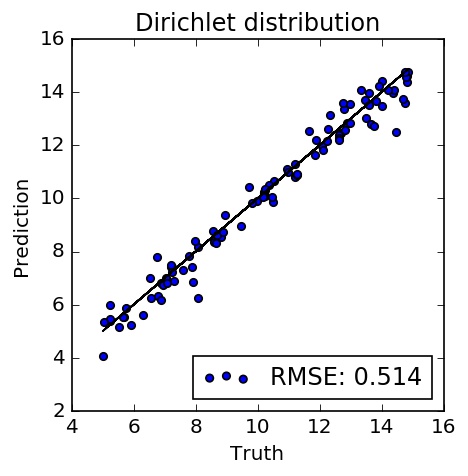} }}
	\subfigure[]{{		\includegraphics[width=4.5cm, height=4.5cm]{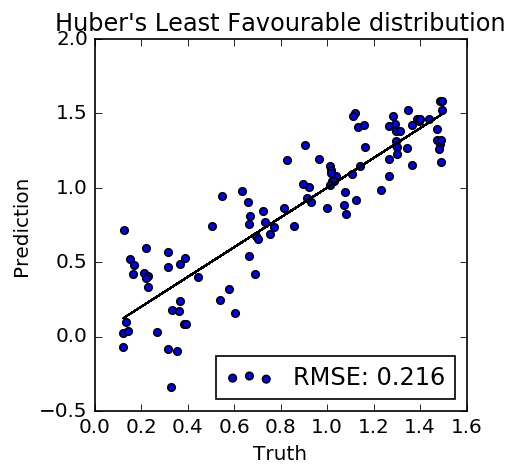} }}
	\subfigure[]{{	     \includegraphics[width=4.5cm, height=4.5cm]{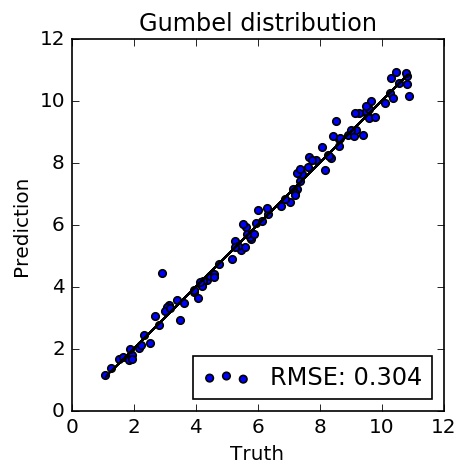} }}
	\subfigure[]{{		\includegraphics[width=4.5cm, height=4.5cm]{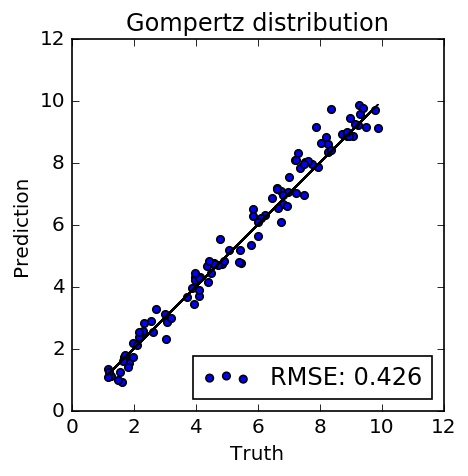} }}
	\subfigure[]{{		\includegraphics[width=4.5cm, height=4.5cm]{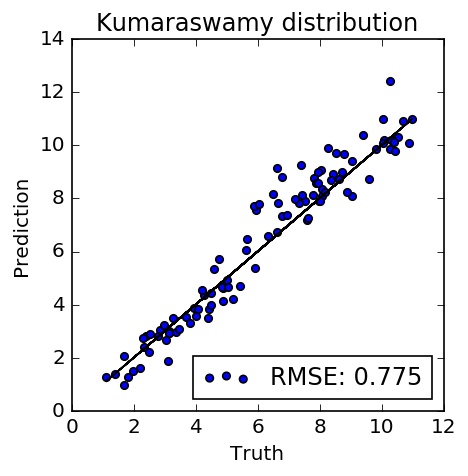} }}
	\subfigure[]{{		\includegraphics[width=4.5cm, height=4.5cm]{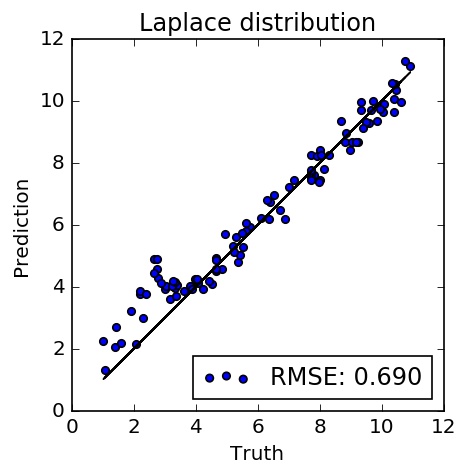} }}
	\subfigure[]{{		\includegraphics[width=4.5cm, height=4.5cm]{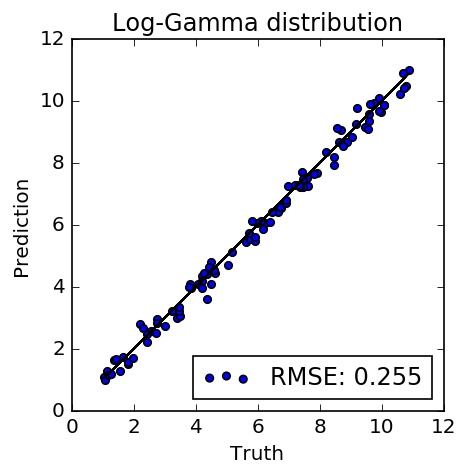} }}
	
	\caption{Distribution wise performance of PSA-5 neural parameter estimator under large CNN on the test dataset with $N=900$. RMSE is reported for each distribution. Part 3.}
	\label{scatterplotpart3}
\end{figure}

\begin{figure}
	\centering
	\subfigure[]{{		\includegraphics[width=4.5cm, height=4.5cm]{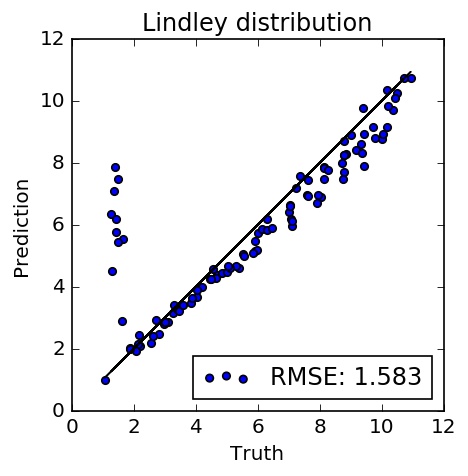} }}
	\subfigure[]{{		\includegraphics[width=4.5cm, height=4.5cm]{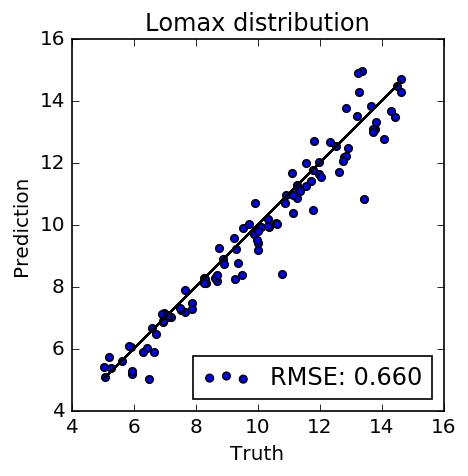} }}
	\subfigure[]{{		\includegraphics[width=4.5cm, height=4.5cm]{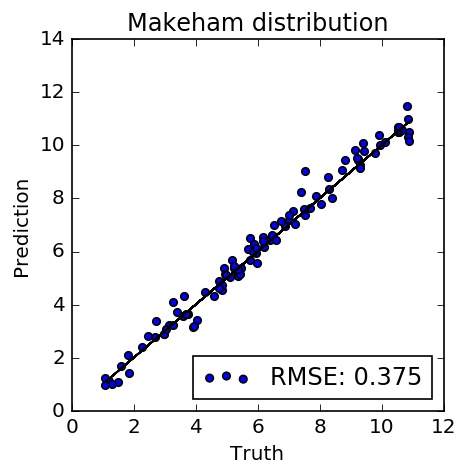} }}
	\subfigure[]{{		\includegraphics[width=4.5cm, height=4.5cm]{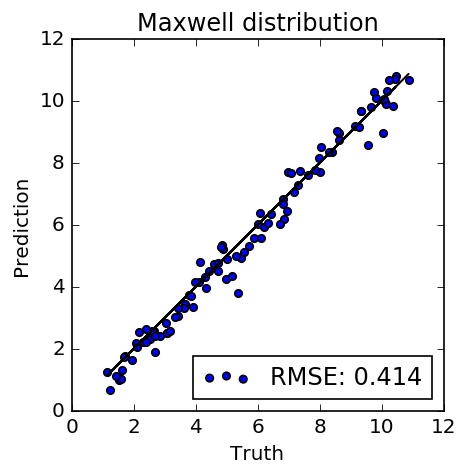} }}
	\subfigure[]{{		\includegraphics[width=4.5cm, height=4.5cm]{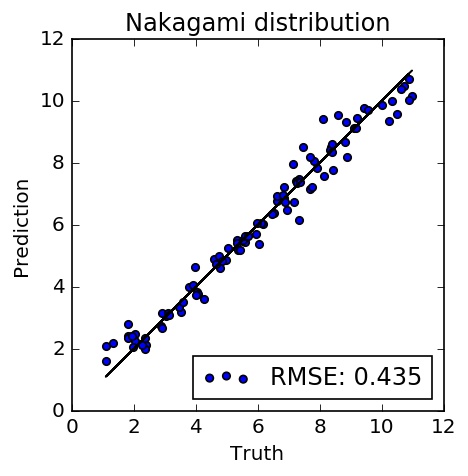} }}
	\subfigure[]{{		\includegraphics[width=4.5cm, height=4.5cm]{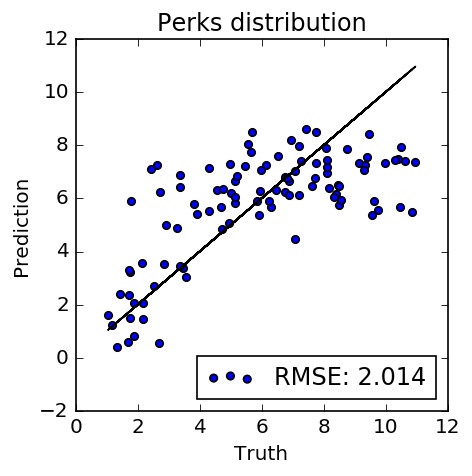} }}
	\subfigure[]{{		\includegraphics[width=4.5cm, height=4.5cm]{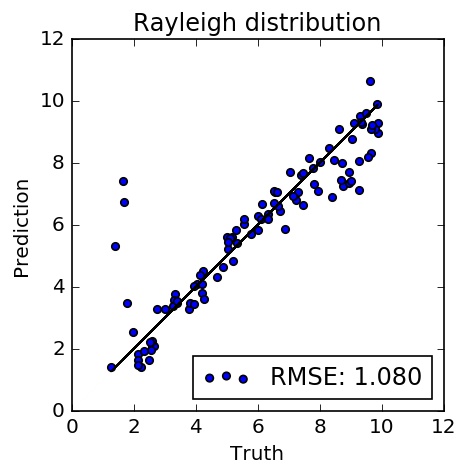} }}
	\subfigure[]{{		\includegraphics[width=4.5cm, height=4.5cm]{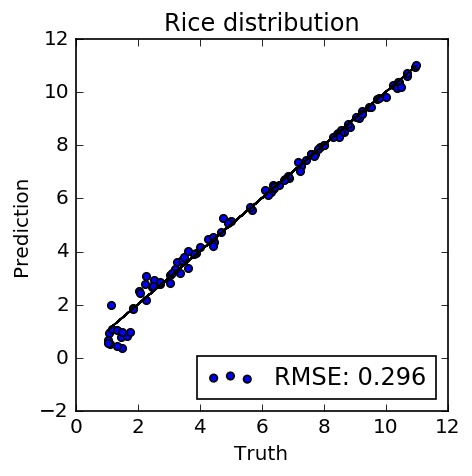} }}
	\subfigure[]{{		\includegraphics[width=4.5cm, height=4.5cm]{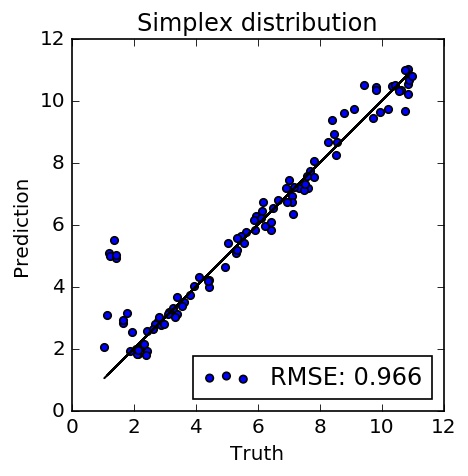} }}
	\subfigure[]{{		\includegraphics[width=4.5cm, height=4.5cm]{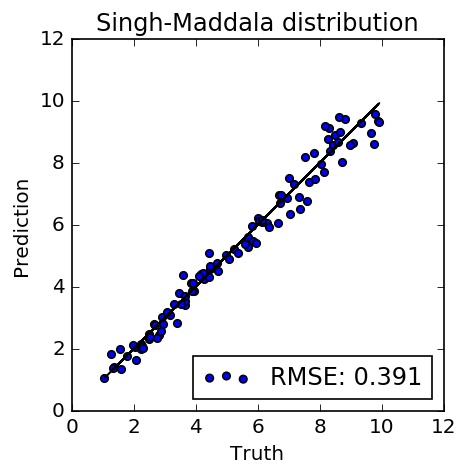} }}
	\subfigure[]{{		\includegraphics[width=4.5cm, height=4.5cm]{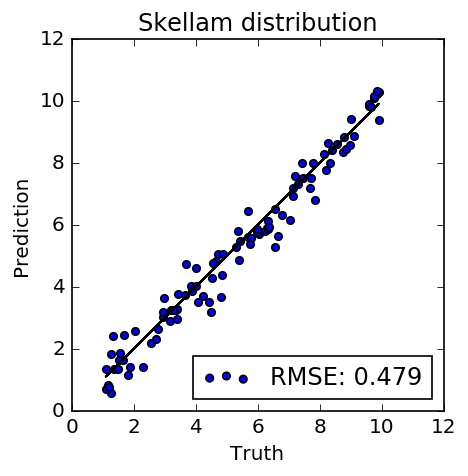} }}
	\subfigure[]{{		\includegraphics[width=4.5cm, height=4.5cm]{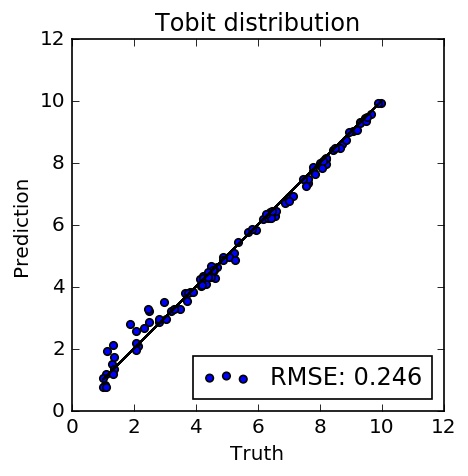} }}

	\caption{Distribution wise performance of PSA-5 neural parameter estimator under large CNN on the test dataset with $N=900$. RMSE is reported for each distribution. Part 4.}
	\label{scatterplotpart4}
\end{figure}

\begin{figure}
	\centering
	\subfigure[]{{		\includegraphics[width=4.5cm, height=4.5cm]{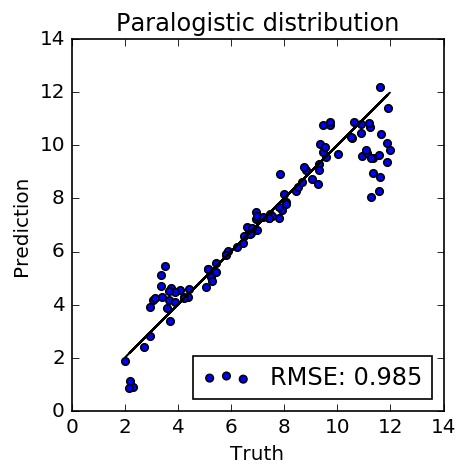} }}
	\subfigure[]{{		\includegraphics[width=4.5cm, height=4.5cm]{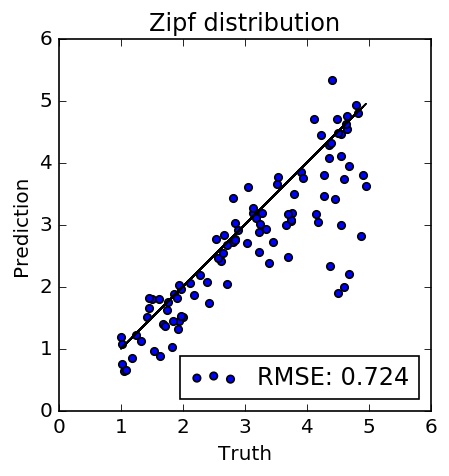} }}
	\caption{Distribution wise performance of PSA-5 neural parameter estimator under large CNN on the test dataset with $N=900$. RMSE is reported for each distribution. Part 5.}
	\label{scatterplotpart5}
\end{figure}

\end{document}